%% file: main.tex
\documentclass[acmsmall,screen]{MapleTrans} 

\usepackage[normalem]{ulem}

\usepackage{enumitem}
\usepackage{etoolbox}
\usepackage{breqn}
\usepackage{lineno}
\BeforeBeginEnvironment{dmath}{\begin{nolinenumbers}}%
\AfterEndEnvironment{dmath}{\end{nolinenumbers}}
\usepackage{maple}
\usepackage{xfrac}
\usepackage{xcolor}
\definecolor{mygreen}{rgb}{0,0.6,0}
\definecolor{mygray}{rgb}{0.5,0.5,0.5}
\definecolor{mymauve}{rgb}{0.58,0,0.82}
\definecolor{altblue}{rgb}{0.0,0.6,1.0}
\definecolor{lstbg}{cmyk}{0.05, 0.01, 0, 0}
\definecolor{morebluish}{cmyk}{0.06,0.04,0,0}
\usepackage{listings}
\input{listings-maple-definition.sty}
\setcopyright{acmlicensed}
\copyrightyear{2024}
\acmYear{2024}
\acmDOI{XXXXXXXXX}

\acmJournal{MAPLETRANS}
\acmVolume{X}
\acmNumber{Y}
\acmArticle{Z}
\acmMonth{0}

\input{packages/math}
\input{packages/add_ons}
\input{packages/draw}

\begin{document}

\title[On the Certification of the Kinematics of 3-DOF SPMs]{On the Certification of the Kinematics of 3-DOF Spherical Parallel Manipulators}

\author{Alexandre Lê}
\email{alexandre-thanh.le@safrangroup.com}
\email{alexandre.le@inria.fr}
\affiliation{%
  \institution{Safran Electronics \& Defense}
  \streetaddress{100 avenue de Paris}
  \city{Massy}
  \state{Île-de-France}
  \country{France}
  \postcode{91344}
}
\affiliation{%
  \institution{Sorbonne Université, Université de Paris Cité, Institut de Mathématiques de Jussieu Paris Rive Gauche}
  \streetaddress{4 place Jussieu}
  \city{Paris}
  \state{Île-de-France}
  \country{France}
  \postcode{75252 CEDEX 05}
}
\affiliation{%
  \institution{Inria Paris}
  \streetaddress{2 rue Simone Iff}
  \city{Paris}
  \state{Île-de-France}
  \country{France}
  \postcode{75012}
}

\author{Damien Chablat}
\email{damien.chablat@cnrs.fr}
\affiliation{%
  \institution{LS2N, UMR CNRS}
  \city{Nantes}
  \country{France}
}

\author{Guillaume Rance}
\email{guillaume.rance@safrangroup.com}
\affiliation{%
  \institution{Safran Electronics \& Defense}
  \streetaddress{100 avenue de Paris}
  \city{Massy}
  \state{Île-de-France}
  \country{France}
  \postcode{91344}
}

\author{Fabrice Rouillier}
\email{fabrice.rouillier@inria.fr}
\affiliation{%
  \institution{Sorbonne Université, Université de Paris Cité, CNRS, Institut de Mathématiques de Jussieu Paris Rive Gauche}
  \streetaddress{4 place Jussieu}
  \city{Paris}
  \state{Île-de-France}
  \country{France}
  \postcode{75252 CEDEX 05}
}
\affiliation{%
  \institution{Inria Paris}
  \streetaddress{2 rue Simone Iff}
  \city{Paris}
  \state{Île-de-France}
  \country{France}
  \postcode{75012}
}

\renewcommand{\shortauthors}{Lê et al.}

\begin{abstract}
\textbf{Abstract}. This paper aims to study a specific kind of parallel robot: \textit{Spherical Parallel Manipulators} (SPM) that are capable of unlimited rolling. A focus is made on the kinematics of such mechanisms, especially taking into account uncertainties (\textit{e.g.} on conception \& fabrication parameters, measures) and their propagations. Such considerations are crucial if we want to control our robot correctly without any undesirable behavior in its workspace (\textit{e.g.} effects of singularities). In this paper, we will consider two different approaches to study the kinematics and the singularities of the robot of interest: symbolic and semi-numerical. By doing so, we can compute a singularity-free zone in the work- and joint spaces, considering given uncertainties on the parameters. In this zone, we can use any control law to inertially stabilize the upper platform of the robot.
\end{abstract}



\maketitle


\newpage
\section{Context of the study}

	\subsection{Introduction}

In order to take a panorama picture on a moving career using high definition cameras, a classical approach is to use a gimbal system \cite{Hil08, Mas08}. Gimbal systems can be regarded as serial robots and provide up to 3 DOF (yaw, pitch, roll). However, those ones are limited by their architectures: stabilizing the camera means stabilizing a mass which downgrades the quality of inertial stabilization. If we want to improve the inertial stablization of such devices (in terms of quality and even DOF), a solution can found by studying other architectures: parallel robots.

	\subsection{Generalities on parallel robots}
	
\begin{figure}[htbp]
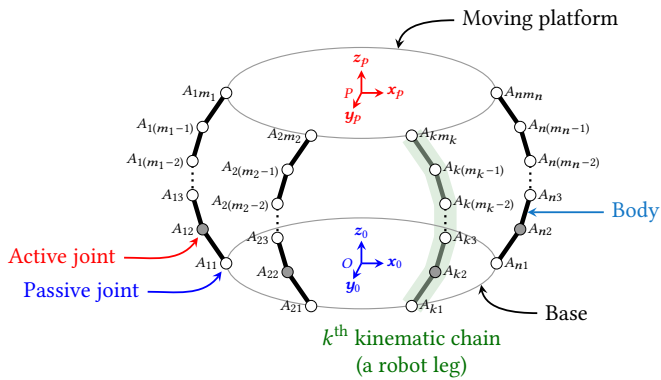

	\centering
	\includestandalone[scale=0.85]{TikZ/robot_parallele_structure}
	\caption{General structure of a parallel robot}
	\label{fig:robot_parallelle_structure}
\end{figure}

	\emph{Parallel robots} are defined in \cite{Lei91} as robots that control the motion of their end-effectors by means of at least two kinematic chains going from the end-effector towards the fixed base. In other words, parallel robots are manipulators that are composed of two plateforms: one at the base and the other at the top called the \emph{moving platform}. These platforms are connected by $n$ kinematic chains that can be regarded as robot legs. Each kinematic chain has joints $A_{ij}$ that can be either motorized with actuactors (we call them \emph{active joints}) -- or not (we call them \emph{passive joints}). Finally, two consecutive joints are linked by a \emph{body}. A typical kinematic chain of a parallel robot is depicted in Figure \ref{fig:robot_parallelle_structure}.

Due to their architecture, parallel robots are mechanisms presenting very good performance in terms of dynamics, stiffness and accuracy to manipulate large loads.	Moreover, such architectures also make it possible to reduce the mass of the movable links. Indeed, all the actuators are mainly fixed on the base and many parts are subject to traction/compression constraints to the extent that it possible to use less powerful actuators. Such nice properties are very suitable for our applications. First appearance of parallel robots are hexapods (in the middle of the 20\textsuperscript{th} century) that are basically used for flight simulations or pneumatic testing using their prismatic legs. However, as they were less widespread than their serial counterparts, studies and knowledge about their modeling had been limited even if they are gaining interests in the recent years (medical, food industry, etc.). 
	
	\subsection{Generalities on SPMs}
	
\begin{figure}[htbp]
	\centering
    \begin{subfigure}{0.35\textwidth}
        \centering
        \includegraphics[scale=0.3]{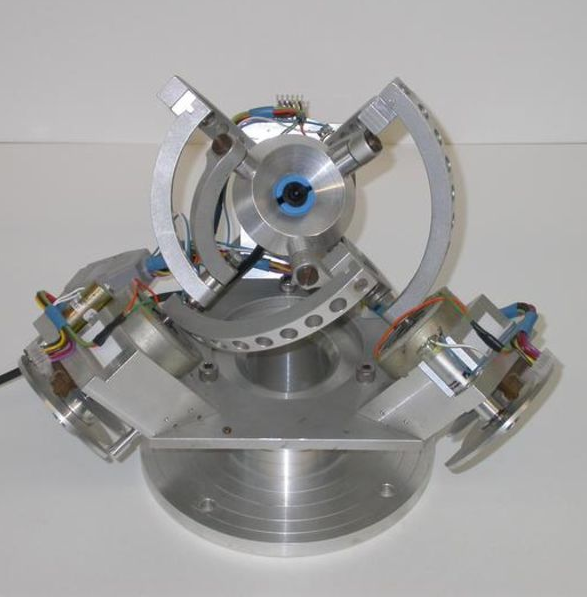}
        \caption{Agile eye}
        \label{fig:ag_eye}
    \end{subfigure}
    \begin{subfigure}{0.35\textwidth}
        \centering
        \includegraphics[scale=0.4]{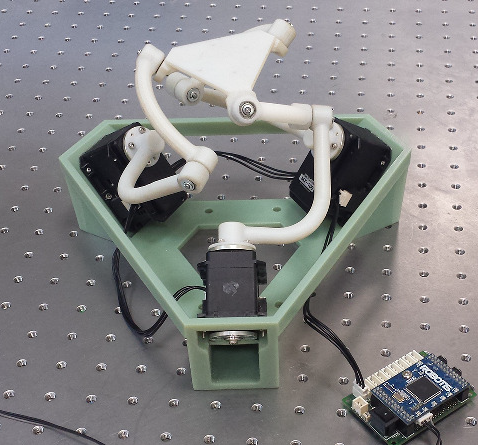}
        \caption{Agile wrist}
        \label{fig:pa_kz}
    \end{subfigure}
    \hspace*{\fill} 
    \begin{subfigure}{0.27\textwidth}
        \centering
        \includegraphics[scale=0.046]{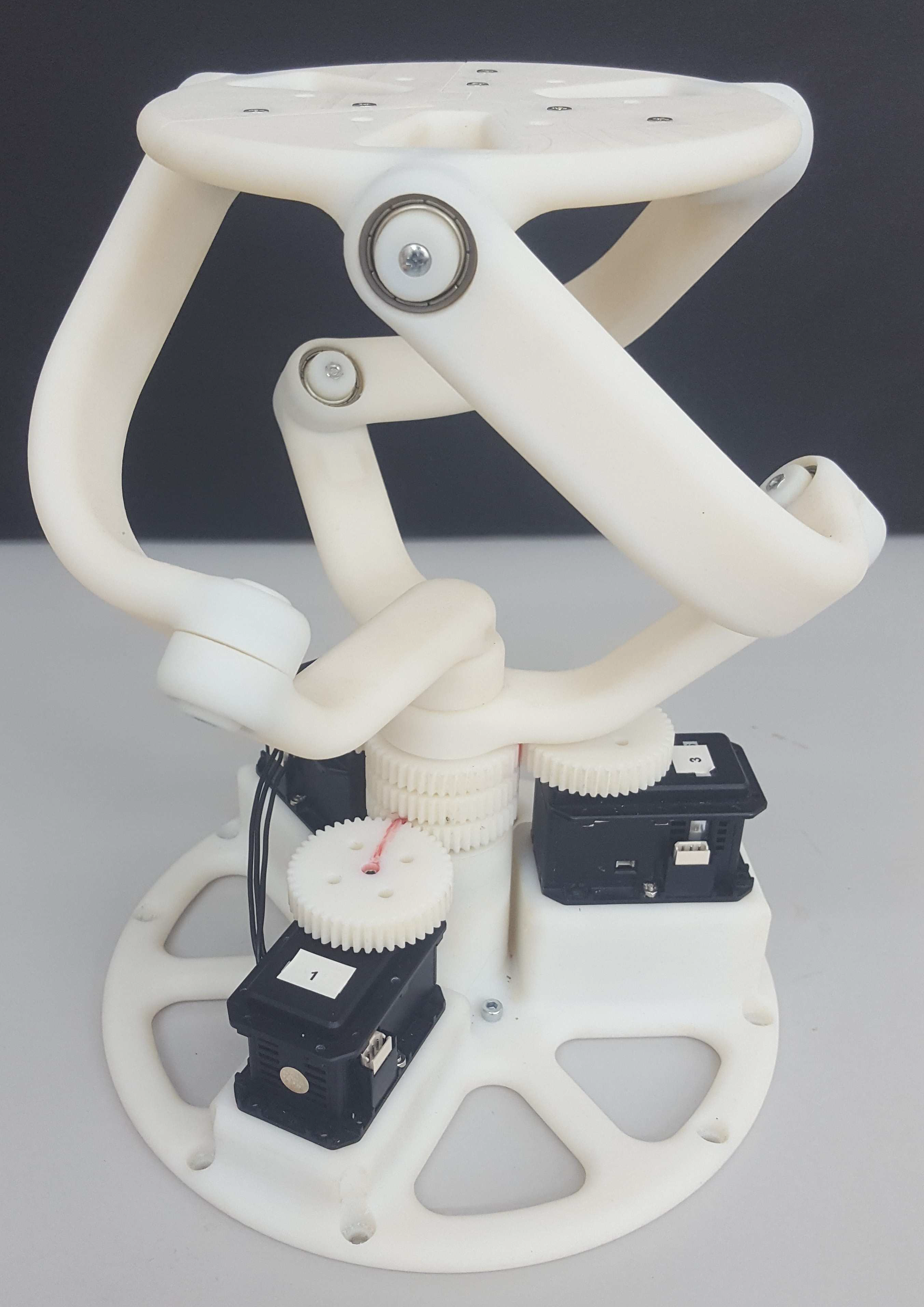}
        \caption{Coaxial agile wrist}
        \label{fig:pa_co}
    \end{subfigure}
	\caption{Examples of non-redundant SPMs (3-RRR)}
	\label{fig:spm_pres}
\end{figure}

There is no unique way \cite{Mer06,KhaBri15} to classify parallel robots (some authors speak in terms of the number of DOF while others focus on their types, \textit{e.g.} revolute or prismatic joints). In this paper, we will especially study a specific type of parallel robots: \emph{Spherical Parallel Manipulators} (SPM) that are non-redundant\footnote{the number of actuators corresponds to the DOF}. SPMs only make rotational motions with their revolute joints. In the special case of non-redundant SPMs, the mobile platform has 3 DOF that we will call \emph{orientation}. The following figures illustrate some remarkable (non-redundant) SPMs:
	
\begin{itemize}
	\item \textsc{Fig}. \ref{fig:ag_eye} depicts the \emph{agile eye} by \cite{GH94} ;
	\item \textsc{Fig}. \ref{fig:pa_kz} depicts the \emph{agile wrist} by \cite{SNR15} ;
	\item \textsc{Fig}. \ref{fig:pa_co} depicts the \emph{agile wrist with coaxial input shafts} by \cite{TNS19}.
\end{itemize}

The last type of SPM is particularly suitable for our case since it allows unlimited rolling that can be useful to obtain a panorama while stabilizing the upper platform in the same time. However, all the non-redundant SPMs have the same modeling that will be detailled in the next section.

\section{Modeling of a non-redundant 3-DOF SPM}

	\subsection{Description}

\begin{figure}[htbp]
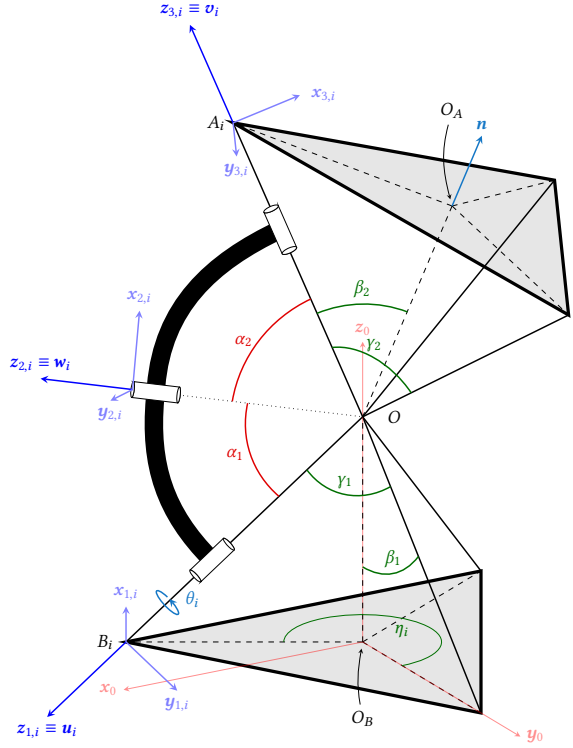

	\centering
	\includestandalone[scale=0.7]{TikZ/poignet_agile_sc_rep_with_etas}
	\caption{Illustration of a typical SPM with conception paramaters (red $+$ green), local frames (dark blue)}
	\label{fig:illustration_spm}
\end{figure}

Before modeling the SPM, let us describe the robot in terms of conception parameters and frames. Figure \ref{fig:illustration_spm} illustrates a typical SPM where these elements are shown. According to this figure, any SPM can be described as a manipulator that has two platforms connected with 3 legs. Each leg has 2 links (or body) and an actuated joint at its base. The actuated joint variables (angles) will be denoted as $\theta_i$ for the $i$\textsuperscript{th} leg. This figure also highlights the fact that SPMs only make pure rotations around $O$ called \emph{center of rotation} of the SPM. Using this property, their motions can be described with only vectors. Those vectors must be expressed in the same frame. For convinience, we will express all the vectors and coordinates in the reference frame $\mathcal{F}_0\bydef\left(O,\bm{x}_0,\bm{y}_0,\bm{z}_0\right)$.

There are three types of vectors: the ones describing the base denoted as $\bm{u}_i$, the ones describing the moving platform denoted as $\bm{v}_i$ and the ones describing the intermediate joints denoted as $\bm{w}_i$, with $i\in\itvd{1,3}$. All these vectors are concurrent in $O$.

First $\rep{\bm{u}_i}{0}$ (\textit{i.e.} $\bm{u}_i$ w.r.t. $\mathcal{F}_0$) can be obtained using the following transformations:

\begin{equation}
	\begin{aligned}
		\rep{\bm{u}_i}{0} &= \bm{R}_z{\left( \eta_i \right)}\,\bm{R}_x{\left( \beta_1-\pi \right)}\,\rep{\bm{z}_0}{0}\\
		&= \mat{-\sin\left(\eta_i\right)\sin\left(\beta_1\right)\\
		   	\cos\left(\eta_i\right)\sin\left(\beta_1\right)\\
			-\cos\left(\beta_1\right)
			}			
	\end{aligned}
\end{equation}

where $\bm{R}_x$ (\textit{resp.} $\bm{R}_y$ and $\bm{R}_z$) denotes the rotation matrix around the local $x$-axis (\textit{resp.} $y$- and $z$-axis). Then, vectors $\bm{w}_i$ w.r.t. frame $\mathcal{F}_0$ can be expressed as:
\begin{equation}
	\begin{aligned}
		\rep{\bm{w}_i}{0} &= \bm{R}_z{\left( \eta_i \right)}\,\bm{R}_x{\left( \beta_1-\pi \right)}\,\bm{R}_z{\left(\theta_i\right)}\,\bm{R}_x{\left( \alpha_1 \right)}\,\rep{\bm{z}_0}{0}\\
			&= \mat{
					-\sin{\left( \eta_i \right)}\sin{\left( \beta_1 \right)}\cos{\left( \alpha_1 \right)} + \sin{\left( \alpha_1 \right)}\left[ \cos{\left( \eta_i \right)}\sin{\left( \theta_i \right)}-\sin{\left( \eta_i \right)}\cos{\left( \beta_1 \right)}\cos{\left( \theta_i \right)} \right]\\
					\cos{\left( \eta_i \right)}\sin{\left( \beta_1 \right)}\cos{\left( \alpha_1 \right)} + \sin{\left( \alpha_1 \right)}\left[ \sin{\left( \eta_i \right)}\sin{\left( \theta_i \right)}+\cos{\left( \eta_i \right)}\cos{\left( \beta_1 \right)}\cos{\left( \theta_i \right)} \right]\\
					\sin{\left( \beta_1 \right)}\cos{\left( \theta_i \right)}\sin{\left( \alpha_1 \right)}-\cos{\left( \alpha_1 \right)}\cos{\left( \beta_1 \right)}
	}
	\end{aligned}
\end{equation}

Finally, the moving platform vectors $\rep{\bm{v}_i}{0}$ are:
\begin{equation}
\rep{\bm{v}_i}{0} = \bm{M}\,\bm{R}_z{\left( \eta_{i} \right)}\,\bm{R}_x{\left( -\beta_2 \right)}\,\rep{\bm{z}_0}{0}
\end{equation}
where $\bm{M}$ denotes the orientation matrix of the mobile platform and can be expressed using several formalisms (Euler angles, Tait-Bryan angles, quaternions, ...). In our case, the $ZXY$ Tait-Bryan angles are used to describe our orientation:
\begin{equation}\label{eq:rot_conv}
	\bm{M} \bydef \bm{R}_z{\left( \chi_3 \right)}\,\bm{R}_x{\left( \chi_1 \right)}\,\bm{R}_y{\left( \chi_2 \right)}
\end{equation}
These equations highlight the fact that a robot can be described in terms of conception parameters and 2 types of variables: either its joint variables or its end-effector coordinates, namely its moving platform's orientation. The joint variables are real values that belong to the \emph{joint space} $\mathcal{Q}$ and will be put into a vector $\bm{\theta}\bydef\mat{\theta_1&\theta_2&\theta_3}^{\mathsf{T}}$. The end-effector coordinates are real values that belong to the \emph{workspace} $\mathcal{W}$ and will be concatenated into a vector $\bm{\chi}\bydef\mat{\chi_1&\chi_2&\chi_3}^{\mathsf{T}}$. This is fundamental to establish any modeling of a robot.  In this paper, the \emph{kinematics} of SPMs are studied through their \emph{geometric} and \emph{first order kinematic models}.

	\subsection{Geometric and first order kinematic models}

The \emph{geometric model} of a parallel manipulator is a system of equations describing the relationships between the actuated joint variables $\bm{\theta}$ and the coordinates (orientations) $\bm{\chi}$ of the moving platform. 

\begin{figure}[htbp]
    \centering
    \begin{tikzpicture}[>=latex',rounded corners]
        \draw[draw=none,fill=bleu_identitaire!30,fill opacity=0.25,text opacity=1] (-1,-2) rectangle (1,2);
        \draw[draw=none] (-1,-2)--(-1,2)node[midway,left=0.5em,bleu_identitaire]{$\mathcal{W}$};
        \node at (0,0.75)(dotx) {$\bm{\chi}$};
        
        \begin{scope}[xshift=15em]
        	\draw[draw=none,fill=dred!30,fill opacity=0.25,text opacity=1] (-1,-2) rectangle (1,2);
        	\draw[draw=none] (1,2)--(1,-2)node[midway,right=0.5em,dred]{$\mathcal{Q}$};
        	\node at (0,0.75)(dqa) {$\bm{\theta}$};
        	\node at (0,-0.75)(dqd) {$\bm{\theta}_d$};
        \end{scope}
        
        \draw[thick,->,bleu_identitaire] (dotx) to[in=150,out=30] node[midway,fill=bleu_identitaire!5,fill opacity=0.95,text opacity=1]{Inverse problem} (dqa);
        \draw[thick,->,bleu_identitaire,dashed] (dqa) -- (dqd);
        \draw[thick,->,dred] (dqa) to[in=0,out=180] node[midway,fill=dred!5,fill opacity=0.95,text opacity=1]{Forward problem} (dotx);
        \draw[thick,->,bleu_identitaire,dashed] (dotx) to[in=-160,out=-80] (dqd);
    \end{tikzpicture}
    \caption{Principle of the geometric model}
    \label{fig:mg_sb}
\end{figure}
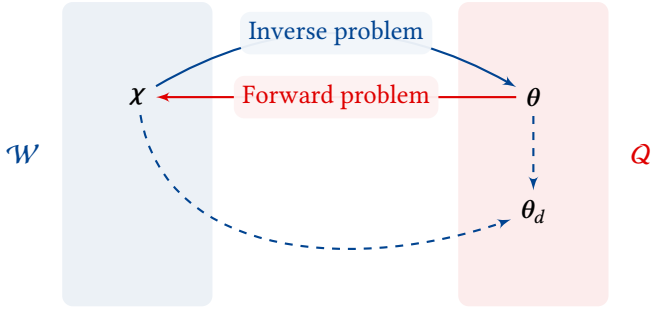

An extended problem is to also consider passive intermediate joints ($\bm{\theta}_d$) which will not be covered in this article. Figure \ref{fig:mg_sb} describes the two points of view of the same problem as previously stated. As we only focus on non-redundant SPMs, their geometric models consist in a system $\bm{f}$ of $n_{\text{dof}}=n_a=3$ independent equations with variables $\bm{\theta}$ and $\bm{\chi}$. The following system describes such a model for SPMs:
\begin{equation}\label{eq:fgm_spm}
\bm{f}\left(\bm{\theta},\bm{\chi}\right) \bydef \mat{\bm{w}_1^{\mathsf{T}}\,\bm{v}_1-\cos{\left(\alpha_2\right)}\\
\bm{w}_2^{\mathsf{T}}\,\bm{v}_2-\cos{\left(\alpha_2\right)}\\
\bm{w}_3^{\mathsf{T}}\,\bm{v}_3-\cos{\left(\alpha_2\right)}} = \bm{0}_{3\times1}
\end{equation}
The forward geometric problem (FGM) is taking \eqref{eq:fgm_spm} with $\bm{\theta}$ being kwown and try to solve it by finding the corresponding $\bm{\chi}$. The solutions found are then called \emph{assembly modes} of the parallel robot. Conversely, the inverse geometric problem (IGM) is taking \eqref{eq:fgm_spm} and try to solve it by finding the corresponding $\bm{\theta}$. The solutions found are then called \emph{working modes} of the parallel robot. By differentiating \eqref{eq:fgm_spm} w.r.t. time, we get the \emph{first order kinematic model} which can be written as $\bm{A}\,\dot{\bm{\chi}}+\bm{B}\,\dot{\bm{\theta}}=\bm{0}_{3\times1}$ where $\bm{A}\bydef\pdv{\bm{f}}{\bm{\chi}}$ denotes the \emph{parallel Jacobian matrix} and $\bm{B}\bydef\pdv{\bm{f}}{\bm{\theta}}$ denotes the \emph{serial Jacobian matrix}.

\section{Implementation of the geometric model}

	\subsection{Requirements and strategy}
	
The geometric model was hitherto obtained through symbolic computation (here using \textsc{Maple} software 2022 version). However, in order to be implemented, we must:
\begin{itemize}
	\item \textbf{define a prescribed workspace $\mathcal{W}^\star$}: in our case we want to ensure that our SPM can stabilize its moving platform up to $\pm20^\circ$ in roll \& pitch. Thus, $\mathcal{W}^\star$ is the set
	\begin{equation}
		\mathcal{W}^\star \bydef \left\{ \left(\chi_1,\chi_2,\chi_3\right)\in\mathbb{R}^3 \text{\;s.t.\;} \abs{\chi_1}\leq 20^\circ\text{\;and\;} \abs{\chi_2}\leq 20^\circ\right\} 
	\end{equation}
	\begin{remark}
		We previously defined $\mathcal{W}$ as the \emph{workspace} of our robot which is the set of all orientations $\bm{\chi}$ that its moving platform can reach. However, we only focus on our \emph{prescribed workspace} $\mathcal{W}^\star\subseteq\mathcal{W}$. This distinction is useful to upgrade/optimize the performances of our robot that are subjects to specifications.
	\end{remark}
	\item \textbf{specify conception parameters} (see Tab. \ref{tab:parameters}): $\beta_1=0$ is chosen in order to have coaxial input shafts (see Fig. \ref{fig:pa_co} and \ref{fig:illustration_spm}). The $\eta_i$s are defined such that the platforms' joints are regularly spaced. The other conception parameters are chosen using the global conditioning index approach and optimal values determined in \cite{Bai10} (see Appendix \ref{sec:a1}).
\begin{table}[htbp]
    \centering
    {\renewcommand{\arraystretch}{1} 
    {\setlength{\tabcolsep}{0.45cm} 
    \begin{tabular}{c|ccccc}
    \hline\hline
    Parameters & $\eta_i$ & $\alpha_1$ & $\alpha_2$ & $\beta_1$ & $\beta_2$\\
    Values (rad) & $2(i-1)\pi/3$ & $\pi/4$ & $\pi/2$ & $0$ & $\pi/2$\\\hline\hline
    \end{tabular}
    }} 
    \caption{Exact values of conception parameters}
    \label{tab:parameters}
\end{table}
	\vspace*{-1em}
	\item be able to \textbf{certify} the SPM's \textbf{FGM} and \textbf{IGM} in order to \textbf{solve} them correctly.
\end{itemize}

The last point is crucial. Although the system $\bm{f}$ is non-linear (which can make the FGM and IGM harder to solve), it can be turned into a polynomial system through appropriate changes of variables. This allows $\bm{f}$ to become a system $\bm{S}$ of the form 
\begin{equation}
	\bm{S}\bydef\left\{p_1(\bm{U},\bm{X})=0,\dots,p_{n_a}(\bm{U},\bm{X})=0\right\}
\end{equation}
where $\bm{U}=(U_1,\dots,U_d)$ is the $d$-uple of parameters, $\bm{X}=(X_1,\dots,X_{n})$ the $n$-uple of unknowns and $p_i$, $i\in\itvd{1,n_a}$ being polynomials in the indeterminates $\bm{U}$, $\bm{X}$ with rational coefficients. The tangent half-angle formulas are interesting changes of variables: this substitution has the advantage to keep the same number of equations, parameters and variables ($n_a=d=n=3$).

\begin{remark}
	This would not be case for sine/cosine changes of variables that double the number of variables and equations. 
\end{remark}

Depending on the point of view (IGM or FGM), $\bm{U}$ and $\bm{X}$ can either be $\bm{j}\bydef\left\{ j_i=\tan{(\theta_i/2)}, i=\itvd{1,3} \right\}$ or $\bm{o}\bydef\left\{ o_i=\tan{(\chi_i/2)}, i=\itvd{1,3} \right\}$. Additionally, such changes of variables are still valid considering our joint- and workspace: $\theta_i,\chi_1,\chi_2\neq\pm\pi\;[2\pi]$. Appendix \ref{sec:a2} shows the explicit expression of $\bm{S}$, the geometric model of our SPM in its polynomial form. We also assume that $\bm{S}$ has a finite number of complex solutions: for almost all $d$-uples $\bm{u}\bydef\left(u_1,\dots,u_d\right)\in\mathbb{C}^d$, the system $\left.\bm{S}\right|_{\bm{U}=\bm{u}}=\left\{ p_1(\bm{u},\bm{X})=0,\dots,p_{n_a}(\bm{u},\bm{X})=0 \right\}$ has finitly many complex solutions. Such a system is called \emph{generically zero-dimensionnal} and will be solved using \emph{Algebraic Geometry} techniques \cite{CLO15} by associating $\bm{S}$ with $\mathcal{I}=\bra{p_1,\dots,p_{n_a}}$ being the \emph{ideal} of $\mathbb{Q}[\bm{U},\bm{X}]$ generated by the polynomials $p_1,\dots,p_{n_a}$, such that $\overline{\proj_{\bm{U}}\left(\mathcal{V}\left(\mathcal{I}\right)\right)}=\mathbb{C}^d$, where $\proj_{\bm{U}}$ denotes the projection onto the parameter space and $\overline{\mathcal{V}}$ the closure of any subset $\mathcal{V}\subset\mathbb{C}^d$. Thus, the complex solutions of $\bm{S}$ define the \emph{algebraic variety} $\mathcal{V}(\mathcal{I})$.

\medskip

However, we also want to go further by considering the robustness of our SPM: solving \eqref{eq:fgm_spm} means assuming that the (theorical) conception parameter values will perfectly correspond to the real case values, which is a strong hypothesis and obviously not true. Indeed, parallel robots are unavoidably subject to uncertainties such as fabrication parameters (assembling tolerances of the mechanical parts) or noise in the sensors. Additionally, implementing the kinematics requires numerical approximations (convert irrational parameter values into rational ones).

\medskip

\begin{figure}[htbp]
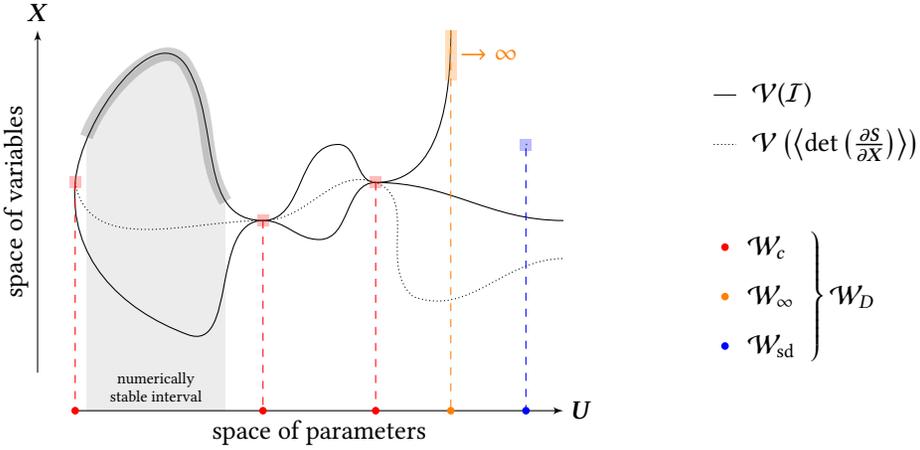

	\centering
	\includestandalone[]{TikZ/var_dis}
	\caption{Certification by avoiding the discriminant variety $\mathcal{W}_D$ w.r.t. the projection onto the paramater space}
	\label{fig:var_dis}
\end{figure}

Despite this, we want to ensure that for small deformations of parameters, solutions found will still be close to the perfect case. In other words, we want to \emph{certify} our SPM's modeling. One of the tools dedicated to this certification work is the notion of \emph{discriminant variety}. This object is closely related to the idea of \emph{projection} onto the space of paramaters $\proj_{\bm{U}}$, as illustrated in Figure \ref{fig:var_dis}. The goal is to have a set of parameters that does not meet and is far from all the numerical unstabilities of the so called discriminant variety. First let us recall its definition from \cite{LazRou07, CMRW20}. 

\begin{definition}[Discriminant Variety]
The discriminant variety of $\mathcal{V}(\mathcal{I})$ w.r.t. $\proj_{\bm{U}}$ denoted as $\mathcal{W}_D$ is the smallest algebraic variety of $\mathbb{C}^d$ such that given any simply connected subset $\mathcal{C}$ of $\mathbb{R}^d\,\backslash\,\mathcal{W}_D$, the number of real solutions of $\bm{S}$ is constant over $\bm{U}$. In our case,
$$
\mathcal{W}_D\bydef \mathcal{W}_{\mathrm{sd}} \cup \mathcal{W}_c \cup \mathcal{W}_\infty
$$
where:
\begin{itemize}
	\item $\mathcal{W}_{\mathrm{sd}}$ is the closure of the projection by $\proj_{\bm{U}}$ of the components of $\mathcal{V}(\mathcal{I})$ of dimension $<d$
	\item $\mathcal{W}_c$ is the union of the closure of the critical values of $\proj_{\bm{U}}$ in restriction to $\mathcal{V}(\mathcal{I})$ and of the projection of singular values of $\mathcal{V}(\mathcal{I})$
	\item $\mathcal{W}_\infty$ is the set of $\bm{U}=\left(U_1,\dots,U_d\right)$ such that $\proj^{-1}\left(\mathcal{C}\right)\cap \mathcal{V}(\mathcal{I})$ is not compact for any compact neighborhood $\mathcal{C}$ of $\bm{U}$ in $\proj_{\bm{U}}\left(\mathcal{V}(\mathcal{I})\right)$.
\end{itemize}
\end{definition}

In our case of non-redundant SPM, we have as many polynomials ($p_1,\dots,p_3$) as unknowns ($X_1,\dots,X_3$) which involves that $\mathcal{W}_{\mathrm{sd}}=\varnothing$. 

	\subsection{Uncertainty analysis}
	
		\subsubsection{Propagation of uncertainty on the fabrication parameters}
	
In order to be \emph{numerically} implemented and from the practical point of view, we need to ensure that the modelling is still equivalent to and valid for a ``deformed'' system (\textit{e.g.} approximation of irrational numbers $\sqrt{2}$, $\sqrt{3}$, expressing the polynomial system $\bm{S}$ with only rational or integer coefficients, uncertainties on fabrication parameters). In particular, it is worth analyzing the impact of such uncertainties on the coefficients of our system. This can be done by considering \emph{interval analysis} tools such as \emph{interval arithmetic} \cite{Neu90,Mer04} or \emph{ball arithmetic} \cite{Hoe10,Joh19} where computations are made with intervals instead of real or float numbers. Both tools allow numerical computations to be more rigorous by taking into account all the possible uncertainties being purely numerical (\textit{e.g.} round-off errors) or physical. In the specific case of ball arithmetic, intervals are rather called \emph{ball intervals}.

\begin{definition}[Ball interval]
	A \emph{ball interval} $[m\pm r]$ is defined as the set of real numbers $x$ such that $x\in\left[m-r,m+r\right]$ where $m$ denotes the \emph{midpoint} of the ball interval and $r$ its \emph{radius}.
\end{definition}

From the computational point of view, $m$ and $r$ are binary floating-point numbers, \textit{i.e.} $m,r\in\mathbb{Z}\,2^{\mathbb{Z}}$, although all the real numbers included in the interval are considered.  This tool is implemented in the \texttt{arb} C library\footnote{see \url{https://arblib.org/}} and is availiable in \textsc{Maple} (v. $\geq 2022$) through the \verb?RealBox(?$m$\verb?,?$r$\verb?)? function. Such a formalism is suitable for a rigourous and reliable computation on the error bounds and will be used in this article to analyze the propagation of fabrication parameters uncertainties on the system of interest. By introducing a realistic uncertainty of $r=10^{-5}\;\text{rad}$ on the fabrication parameters using the \texttt{RealBox} function, the coefficients of $\bm{S}$ (depending on $o_i$ and $j_i$, $i\in\itvd{1,n_a}$) have in the worst case an uncertainty of $r'_{\max}=7\times10^{-5}$. 

		\subsubsection{On the coaxiality of the input shafts}\label{ss:coaxiality}

Another important question deals with the coaxiality of the SPM's input shafts. Indeed, in theory, the actuators must be \emph{perfectly} concentric to allow an illimited rotation around the $z$-axis (yaw). This is nevertheless not the case in practice because of the uncertainties on the fabrication parameters, \textit{i.e.} $\beta_1$ in our modelling is not exactly equal to $0$ for each leg of the SPM. Despite this unfavorable theoretical argument, experimental prototypes \cite{TNS19} have shown that such a mechanism is absolutely capable of moving this way. That leads to say that among all the possible geometrical configurations induced by the uncertainties on $\beta_1$, the coaxial SPM can make an illimited rotation around the yaw axis because of the backlashes of its actuated joints. By undergoing such a phenomenon, the robot can be associated with a virtual one having a perfect axis of coaxiality. Thus, studying the system in its exact form makes sense. Using the above-mentionned approach and given this context, let us certify the IGM and FGM.

\section{Certifying the Inverse Geometric Model of 3-DOF SPMs}

	\subsection{Workspace analysis}

As previously stated, solving the IGM is equivalent to solve $\bm{S}$ with $\bm{o}\equiv\bm{\chi}$ being (orientation) parameters related and (joint) unknowns $\bm{j}\equiv\bm{\theta}$. The goal is to ensure that each orientation value of our prescribed workspace $\mathcal{W}^\star$ has the same number of distinct working modes. In addition, this fact must hold despite data uncertainties such as small variations on parameters $\bm{o}$. However, there is one special case that does not verify those properties and that we want to avoid: numerical unstabilities of the IGM. One of them are \emph{Type-1 singularities} (or \emph{serial singularities}). These phenomena appear when matrix $\bm{B}$ from the kinematic model degenerates: the number of (real) distinct working modes varies and small variations on $\bm{\chi}$ in the neighborhood require huge efforts to move $\bm{\theta}$. Therefore, \emph{certifiying} implies checking if we are ``far enough'' from Type-1 singularities and other numerical unstabilities for all the values of $\bm{\chi}\in\mathcal{W}^\star$. One way to check this singularity is to compute the \emph{discriminant variety} of the IGM (in its polynomial form). This object is convinient to compute since we deal with a parametric system in which each polynomial equation has only 1 variable and is of degree 2 such that $\mathrm{IGM}\equiv\bm{S}=\left\{ p_1\left(j_1,\bm{o}\right)=0, p_2\left(j_2,\bm{o}\right)=0, p_3\left(j_3,\bm{o}\right)=0 \right\}$ where $p_i=a_i\,j_i^2+b_i\,j_i+c_i$ with $i\in\itvd{1,n_a=3}$. In this particular case, studying the discriminant variety $\mathcal{W}_D$ w.r.t. the projection onto the orientation space means computing the \emph{resultant} of each polynomial $p_i$, $\pdv{p_i}{j_i}$ w.r.t. variable $j_i$. Hence,
\begin{equation}\label{eq:wd_igm}
	\begin{aligned}
	\mathcal{W}_D(o_1,o_2,o_3) &= \bigcup\limits_{i=1}^{n_a}\res{\left(p_i,\pdv{p_i}{j_i},j_i\right)}\\
	&= \bigcup\limits_{i=1}^{n_a}-\LC{(p_i)}\discrim\left(p_i,j_i\right)
	\end{aligned}
\end{equation}
where $\discrim\left(p_i,j_i\right)=b_i^2-4a_i c_i$ is the \emph{discriminant} of $p_i$ w.r.t. variable $j_i$ and $\LC(p_i)$ is the \emph{leading coefficient} of $p_i(j_i)$. As illustrated in Figure \ref{fig:var_dis}, the discriminant variety $\mathcal{W}_D$ w.r.t. $\proj_{\bm{o}}$ deals with the partition of the (orientation) parameter space in function of the number of working modes. This amounts to study all numerical unstabilities of the IGM according to \eqref{eq:wd_igm} since:
\begin{itemize}
	\setlength\itemsep{-0.15em}
	\item $\discrim{\left(p_i,j_i\right)}=0$ describe a hypersurface in $(o_1,o_2,o_3)$ where at least two working modes are superposed which is synonym of Type-1 singularity. In this case, there is at least one double root and critical values $(o_1,o_2,o_3)$ verifying that belong to $\mathcal{W}_c$. In our case, we have
	
	\begin{equation}\label{eq:w_c}
	\mathcal{W}_c=\left\{\begin{aligned}
			8 \left(o_{3}^{2}+1\right)^{2} \left(o_{1}^{2}+2 o_{1}-1\right)\left(o_{1}^{2}-2 o_{1}-1\right) &= 0,\\[2ex]
			-32 \left(-o_{1}^{4} o_{2}^{4}+4 \sqrt{3}\, o_{1}^{3} o_{2}^{3}+4 o_{1}^{4} o_{2}^{2}+4 \sqrt{3}\, o_{1}^{3} o_{2}-4 \sqrt{3}\, o_{1} o_{2}^{3} \right.\hspace*{3em}\\\left.-o_{1}^{4}-12 o_{1}^{2} o_{2}^{2}-o_{2}^{4}-4 \sqrt{3}\, o_{1} o_{2}+4 o_{2}^{2}-1\right) \left(o_{3}^{2}+1\right)^{2} &= 0,\\[2ex]
			32 \left(o_{1}^{4} o_{2}^{4}+4 \sqrt{3}\, o_{1}^{3} o_{2}^{3}-4 o_{1}^{4} o_{2}^{2}+4 \sqrt{3}\, o_{1}^{3} o_{2}-4 \sqrt{3}\, o_{1} o_{2}^{3}+o_{1}^{4}+12 o_{1}^{2} o_{2}^{2}\right.\hspace*{3em}\\\left.+o_{2}^{4}-4 \sqrt{3}\, o_{1} o_{2}-4 o_{2}^{2}+1\right) \left(o_{3}^{2}+1\right)^{2} &= 0
	\end{aligned}\right\}
	\end{equation}\smallskip
	
	\item $\LC{\left(p_i\right)}=0$ describe a hypersurface in $(o_1,o_2,o_3)$ where at least one solution goes to infinity. Values $(o_1,o_2,o_3)$ verifying that belong to $\mathcal{W}_\infty$. In our case, we have
	
	\begin{equation}\label{eq:w_infty}
		\mathcal{W}_\infty=\left\{\begin{aligned}
			-\sqrt{2}\, o_{1}^{2} o_{3}^{2}-2 \sqrt{2}\, o_{1} o_{3}^{2}+\sqrt{2}\, o_{1}^{2}+\sqrt{2}\, o_{3}^{2}-2 \sqrt{2}\, o_{1}-\sqrt{2} &= 0,\\[2ex]
			-2 \sqrt{2}+2 \sqrt{2}\, \sqrt{3}\, o_{1}^{2} o_{2}-2 \sqrt{2}\, \sqrt{3}\, o_{1}^{2} o_{3}+2 \sqrt{2}\, \sqrt{3}\, o_{2}^{2} o_{3}-2 \sqrt{2}\, \sqrt{3}\, o_{2} o_{3}^{2}\\
			-2 \sqrt{2}\, o_{1}^{2} o_{2}^{2} o_{3}^{2}+12 \sqrt{2}\, o_{3} o_{1} o_{2}+2 \sqrt{2}\, o_{1} o_{2}^{2} o_{3}^{2}-\sqrt{2}\, o_{1}^{2}+\sqrt{2}\, o_{2}^{2}+2 \sqrt{2}\, o_{3}^{2}+2 \sqrt{2}\, o_{1}\\
			+2 \sqrt{2}\, \sqrt{3}\, o_{1}^{2} o_{2} o_{3}^{2}+2 \sqrt{2}\, \sqrt{3}\, o_{1} o_{2} o_{3}^{2}+2 \sqrt{2}\, o_{1}^{2} o_{2}^{2}+\sqrt{2}\, o_{1}^{2} o_{3}^{2}-\sqrt{2}\, o_{2}^{2} o_{3}^{2}\\
			-2 \sqrt{2}\, \sqrt{3}\, o_{1} o_{2} +2 \sqrt{2}\, o_{1} o_{2}^{2}+2 \sqrt{2}\, o_{1} o_{3}^{2}-2 \sqrt{2}\, \sqrt{3}\, o_{2} &= 0,\\[2ex]
			-2 \sqrt{2}+2 \sqrt{2}\, o_{1} o_{2}^{2} o_{3}^{2}-2 \sqrt{2}\, \sqrt{3}\, o_{1}^{2} o_{2}+2 \sqrt{2}\, \sqrt{3}\, o_{1}^{2} o_{3}-2 \sqrt{2}\, \sqrt{3}\, o_{2}^{2} o_{3}+2 \sqrt{2}\, \sqrt{3}\, o_{2} o_{3}^{2}\\
			+2 \sqrt{2}\, \sqrt{3}\, o_{1} o_{2}-2 \sqrt{2}\, o_{1}^{2} o_{2}^{2} o_{3}^{2}+12 \sqrt{2}\, o_{3} o_{1} o_{2}-\sqrt{2}\, o_{1}^{2}+\sqrt{2}\, o_{2}^{2}+2 \sqrt{2}\, o_{3}^{2}+2 \sqrt{2}\, o_{1}\\
			-2 \sqrt{2}\, \sqrt{3}\, o_{1}^{2} o_{2} o_{3}^{2}-2 \sqrt{2}\, \sqrt{3}\, o_{1} o_{2} o_{3}^{2}+2 \sqrt{2}\, o_{1} o_{2}^{2}+2 \sqrt{2}\, o_{1} o_{3}^{2}+2 \sqrt{2}\, \sqrt{3}\, o_{2}\\
			+2 \sqrt{2}\, o_{1}^{2} o_{2}^{2}+\sqrt{2}\, o_{1}^{2} o_{3}^{2}-\sqrt{2}\, o_{2}^{2} o_{3}^{2} &= 0
		\end{aligned}\right\}
	\end{equation}\smallskip
\end{itemize}

\begin{figure}[htbp]
	\centering
	\begin{subfigure}{0.49\textwidth}
        \centering
        \includegraphics[scale=0.575]{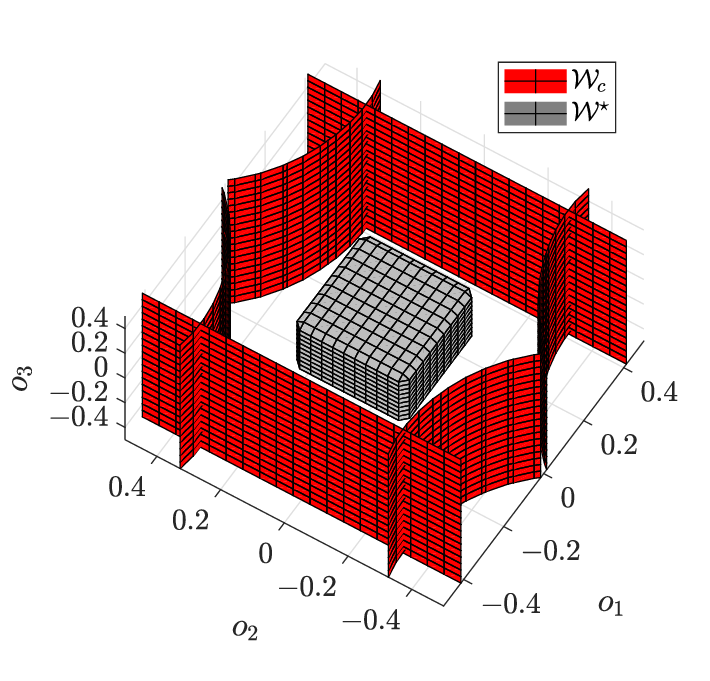}
        \caption{Critical points of the IGM (Type-1 singularities)}
        \label{fig:wc_igm_exact}
    \end{subfigure}
    \begin{subfigure}{0.49\textwidth}
        \centering
        \includegraphics[scale=0.575]{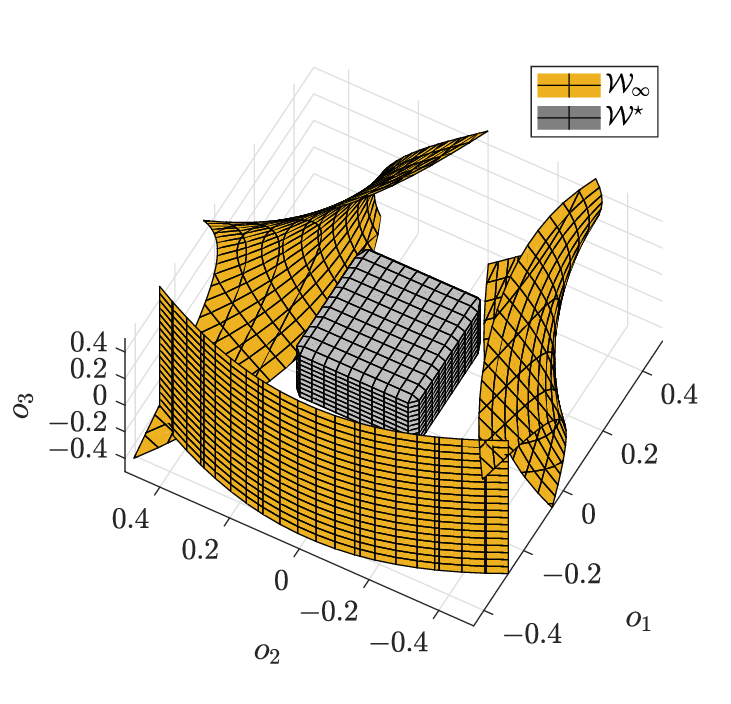}
        \caption{``Infinite points'' of the IGM}
        \label{fig:wi_igm_exact}
    \end{subfigure}
	
	\caption{Discriminant variety of the IGM w.r.t. the projection onto the orientation space}
	\label{fig:wd_igm}
\end{figure}

Both cases imply a drop in the number of working modes. Figure \ref{fig:wd_igm} depicts the plot of the discriminant variety of the SPM's IGM. We can notice that $\mathcal{W}_c$ representing all the Type-1 singularities of our robot is invariant w.r.t. $o_3\equiv\chi_3$. This makes sense since we consider the rotation w.r.t. yaw first (see \eqref{eq:rot_conv}) and the unlimited rolling property allows our robot to yaw without changing its geometry. But most importantly, our workspace $\mathcal{W}^\star$ does not meet the discriminant variety of the IGM ($\mathcal{W}_D$ w.r.t. $\proj_{\bm{o}}$). More precisely, as we deal with 3 quadratic polynomials in $j_i$, solving the IGM for each orientation $\bm{\chi}\in\mathcal{W}^\star$ implies finding $2^3=8$ distinct solutions (working modes). Therefore, we can guarentee that our SPM's IGM will not meet any numerical unstability especially Type-1 singularities, given our conception parameters and our prescribed workspace $\mathcal{W}^\star$: we have certified the IGM of our SPM for our application in its exact form.

\begin{remark}
	From now on, we will only focus on $\mathcal{W}_c$, the set of the critical points of the IGM corresponding to all Type-1 singular configurations. Indeed, $\mathcal{W}_\infty$ is only a concern if we solve the IGM in its current polynomial form whereas $\mathcal{W}_c$ basically depends on our SPM's geometry.
\end{remark}

\begin{figure}[htbp]
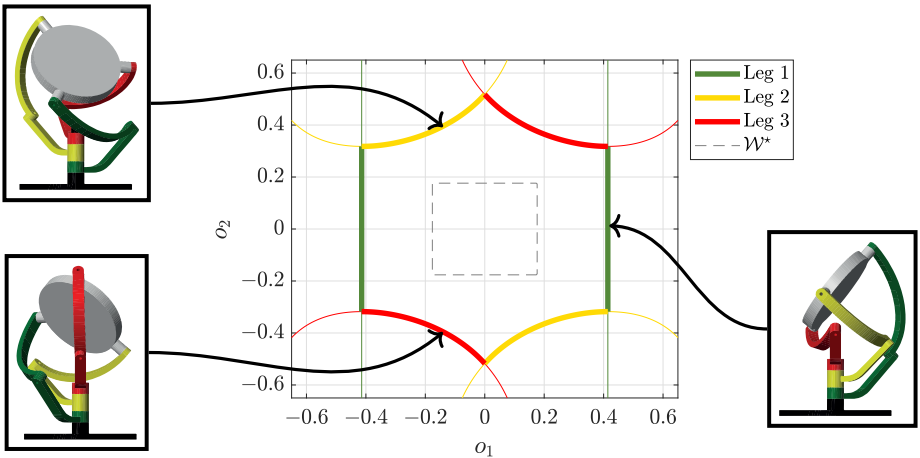

	\centering
	\includestandalone[]{TikZ/pa_co_st1_EN_overview}
	\caption{Type-1 singularity loci of the SPM in the orientation space}
	\label{fig:pa_co_st1}
\end{figure}

Figure \ref{fig:pa_co_st1} depicts the critical values of the IGM in the $(o_1,o_2)$-plan by highlighting some Type-1 singular configurations. The latter confirms the loss of at least 1 DOF as all the configurations have a fully folded or extended leg. A similar (but non-certified) result can be obtained using the conditoning index approach (see Appendix \ref{sec:a3}, Fig. \ref{fig:ci_type1}).

\begin{remark}
	One trivial serial singular configuration can be found by solving the first equation of \eqref{eq:w_c}. We then obtain $o_1=\left\{ -1\pm\sqrt{2},1\pm\sqrt{2} \right\}$ or $\chi_1=\left\{\pm45^\circ,\pm135^\circ\right\}$. The Type-1 singularities of interest comes with any orientation having a roll of $\chi_{1,\mathrm{sing}}=\pm45^\circ$. Starting with $\chi_1=0$, having $\abs{\chi_1}>45^\circ$ is impossible due to the first leg (the green one in Fig. \ref{fig:pa_co_st1}) being totally extended.
\end{remark}

\begin{figure}[htbp]
	\centering
	\includegraphics[scale=0.575]{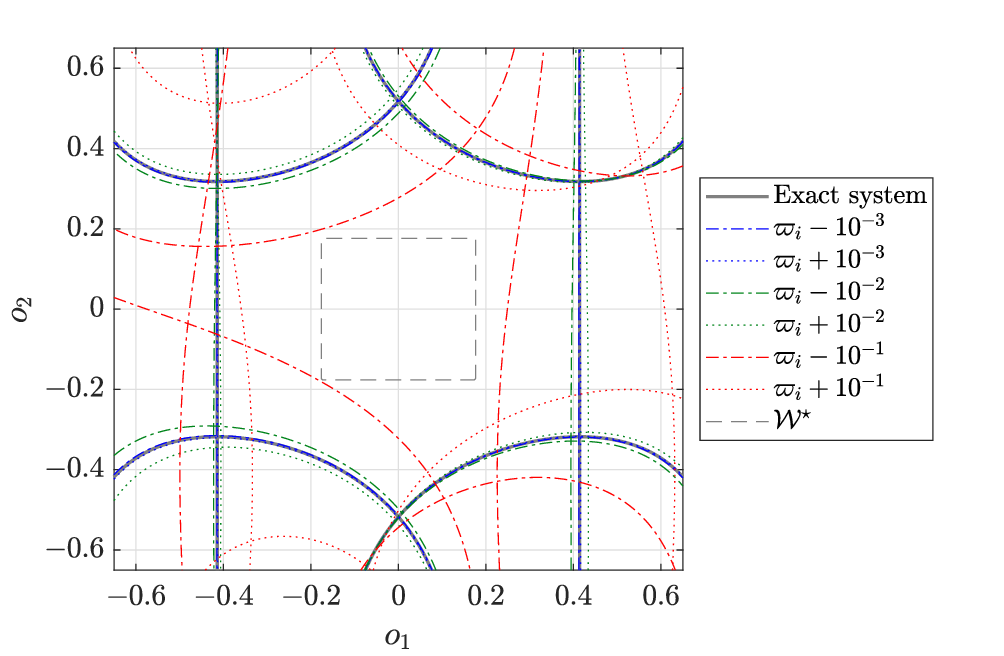}
	\caption{Type-1 singularity loci of the SPM in the orientation space considering uncertainties on the fabrication parameters}
	\label{fig:pa_co_st1_uncertainties}
\end{figure}

It might also be of interest to determine the maximum tolerance on the fabrication parameters towards the Type-1 singularities. In the sequel, we introduce an uncertainty on the fabrication parameters belonging to $\bm{\varpi}$ defined as
\begin{equation}
	\bm{\varpi} \bydef \mat{\alpha_1 & \alpha_2 & \beta_2 & \eta_1 & \eta_2 & \eta_3}^{\mathsf{T}}
\end{equation}
Given our remarks on the coaxiality of our mechanism (see Subsection \ref{ss:coaxiality}), we set $\beta_1=0$ which also keeps the invariance of the IGM's discriminant variety w.r.t. $o_3$. Figure \ref{fig:pa_co_st1_uncertainties} depicts the discriminant variety of the IGM considering such uncertainties.

As seen on this figure, we can introduce an uncertainty up to $10^{-1}\;\text{rad}$ on the fabrication parameters $\varpi_i$ before meeting the discriminant variety of the IGM, which is in fact a comfortable margin.

So far, we have proved that our prescribed workspace $\mathcal{W}^\star$ is Type-1 singular-free. However, from the pratical point of view, it is better to translate this information into the joint space in order to set the actuated joint stops. These limits in $\theta_1$, $\theta_2$ and $\theta_3$ play a double role since they allow our robot to move within $\mathcal{W}^\star$ and avoid singular orientations at the same time. Such a work deals with the \emph{joint space analysis} of our SPM and is presented in the next subsection.
 
	\subsection{Joint space analysis}
	
By taking into account the invariance in orientation w.r.t. yaw, the idea is to set $\chi_3\equiv o_3=0$ and compute a unique working mode from the same leaf of solution for a certain number of pair $(o_1,o_2)$ recovering $\mathcal{W}^\star(o_3=0)$. Such a set is a square in the $(o_1,o_2)$-plane. The uniqueness of the working mode is obtained by choosing an initial joint vector $\bm{\theta}_0$ being one of the working modes corresponding to the SPM's equilibrium orientation, \textit{i.e.} $\chi_{0,i}=0$, $\forall\,i\in\itvd{1,3}$. By solving \eqref{eq:fgm_spm} being the IGM at equilibrium, we have $\theta_{0,i} = \pm\,\pi/2$, $\forall\,i\in\itvd{1,3}$. The existence of $2^3=8$ distinct working modes confirms the regularity of the robot at equilibrium. We arbitrarily choose $\bm{\theta}_0$ with all values $\theta_{0,i}$ being positive such that $\theta_{i}=\frac{-b_i-\sqrt{\discrim{\left(p_i,j_i\right)}}}{2\LC{\left(p_i,j_i\right)}}$, $\forall\,i\in\itvd{1,3}$. Hence, we have
\begin{equation}\label{eq:eq_config}
	\bm{\chi}_0=\bm{o}_0=\mat{0 & 0 & 0}^{\mathsf{T}} \;\xleftrightarrow{(+++)}\; \left\{\begin{aligned}
		\bm{\theta}_0&=\mat{\pi/2 & \pi/2 & \pi/2}^{\mathsf{T}}\\
		\bm{j}_0 &= \mat{1 & 1 & 1}^{\mathsf{T}}
	\end{aligned}\right.
\end{equation}
All the values of $o_1\equiv\chi_1$ and $o_2\equiv\chi_2$ belonging to $\mathcal{W}^\star(o_3=0)$ are considered in the computation of the IGM even though the square is discretized using \emph{ball intervals}. After paving the whole prescribed workspace with $35\times35=1225$ ball intervals, we compute the IGM for each ball interval by setting $\left[m\left(o_i\right)\pm r\left(o_i\right)\right]$ such that $-\tan{\left(\frac{\pi}{18}\right)}\simeq-\frac{177}{1000}\leq m\left(o_i\right)\leq\frac{177}{1000}\simeq\tan{\left(\frac{\pi}{18}\right)}$ with $r\left(o_i\right)=\frac{5}{900}$, $\forall\,i\in\itvd{1,2}$ and $m\left(o_3\right)=0$ with $r\left(o_3\right)=10^{-4}$. The choice of the radii $r\left(o_1\right)$ and $r\left(o_2\right)$ ensures that the ball intervalls overlap in order to cover the whole set $\mathcal{W}^\star\left(o_3=0\right)$. The obtained results are also expressed with the same formalism as the input data, \textit{i.e.} $\left[m\left(j_i\right)\pm r\left(j_i\right)\right]$, $\forall\,i\in\itvd{1,3}$. Each joint variable $j_i\equiv\theta_i$ has a minimum value and a maximum value such that
\begin{equation}
	\min{\left(m\left(j_i\right)-r\left(j_i\right)\right)} \leq j_i \leq \max{\left(m\left(j_i\right)+r\left(j_i\right)\right)}, \qquad\forall\,i\in\itvd{1,3}
\end{equation}
Those values respectively define the lower and upper bound for the joint stops as shown in Table \ref{tab:joint_stops}.

\begin{table}[H]
	\centering
	\begin{tabular}{cccccc}
		\hline
		Joint $i$	& $\min{\left(j_i\right)}$	& $\max{\left(j_i\right)}$	& Joint stops & $\max{\left(r\left(j_i\right)\right)}$	& $\min{\left(\Delta{\left(p_i,j_i\right)}\right)}$\\\hline\hline
		$1$ 		& $0.6693723886$	& $1.525710784$	& $\theta_1\in\left[67^\circ,114^\circ\right]$ & $0.05831109206017$	& $3.149730917$\\\hline
		$2$ 		& $0.6089969554$	& $2.127382005$	& $\theta_2\in\left[62^\circ,130^\circ\right]$ & $0.18036268138497$	& $10.08465368$\\\hline
		$3$ 		& $0.4729818360$	& $1.702299683$	& $\theta_3\in\left[50^\circ,120^\circ\right]$ & $0.15685467160577$	& $10.01625750$\\\hline
	\end{tabular}
	\caption{Extrema joint values obtained after the computation of the IGM of $\mathcal{W}^\star\left(o_3=0\right)$}
	\label{tab:joint_stops}
\end{table}

By considering the unlimitted rolling of our SPM, \textit{i.e.} $\chi_3\in\mathbb{R}$, we can define $\mathcal{Q}_0^\star$ such that
\begin{equation}
	\mathcal{Q}_0^\star\bydef\left\{\left(\theta_1,\theta_2,\theta_3\right)\in\mathbb{R}^3 \;\middle|\; \begin{aligned}
		67^\circ \leq \theta_1-\chi_3 \leq 114^\circ\\
		62^\circ \leq \theta_2-\chi_3 \leq 130^\circ\\
		50^\circ \leq \theta_3-\chi_3 \leq 120^\circ
	\end{aligned},\quad\forall\,\chi_3\in\mathbb{R}\right\}
\end{equation}

Given our leaf of solution, the image of $\mathcal{W}^\star$ through the IGM is the set $\mathcal{Q}^\star$ defined as
\begin{equation}
	\mathcal{Q}^\star \bydef \left\{ \bm{\theta}=\mat{\theta_1 & \theta_2 & \theta_3}^{\mathsf{T}}\in\mathbb{R}^3 \;\middle|\; \bm{\theta}\in\mathcal{Q}_0^\star \;\text{and}\; \mathrm{FGM}\left(\bm{\theta}\right)\in\mathcal{W}^\star \right\}
\end{equation}
\begin{remark}
	We necessarily have $\mathcal{Q}_0^\star \supset \mathcal{Q}^\star\bydef\mathrm{IGM}{\left( \mathcal{W}^\star \right)}$.
\end{remark}

\section{Certifying the Forward Geometric Model of 3-DOF SPMs}

	\subsection{Issue and adopted strategy}

The goal of the FGM certification is to ensure that the number of assembly modes stays constant for any acceptable joint reference value $\bm{\theta}\in\mathcal{Q}^\star$ allowing the SPM to move within our prescribed workspace $\mathcal{W}^\star$. Moreover, the previous fact must stay true despite the above-mentioned data uncertainties. Special cases that do not verify such conditions are numerical unstabilities including \emph{Type-2 singularities} (or \emph{parallel singularities}). These phenomena appear when matrix $\bm{A}$ from the kinematic model degenerates: the number of distinct assembly modes changes and small variations on $\bm{\theta}$ in the neighborhood implies huge variations on $\bm{\chi}$. The robot loses its rigidity by gaining one (or more) uncontrollable motion: the upper platform can move without any input joint efforts \cite{KhaBri15}. Consequently, such configurations should be avoided: this leads to ensure that the set $\mathcal{Q}^\star$ is non-singular. Certification using the discriminant variety done in the previous section could also theorically be extended to the FGM. In this case, roles between parameters and variables would be switched. However, we would obtain a parametric system in which each equation depends of $o_1$, $o_2$ and $o_3$ at the same time. The discriminant variety of the FGM is thus too substantial to compute. We need to investigate numerical stability and robustness using another approach. One way to ensure such properties is to prove the regularity of the FGM for any $\bm{\theta}\in\mathcal{Q}^\star$ given our application, \textit{i.e.} each $\bm{\theta}\in\mathcal{Q}^\star$ has a \emph{unique} assembly mode $\bm{\chi}$ given the leaf of solution of interest. This will be done considering the \emph{path tracking} problem in orientation.

	\subsection{Path tracking in orientation}
	
A closely related problem to the FGM is the path tracking problem. In our case, the upper platform moves with respect to the base frame. Knowing the joint values, the calculator needs to compute the orientation (FGM) at each step given the sampling rate. This computation can be done using a \emph{Newton iterative scheme}. Such a numerical method estimates the pose of the robot by taking advantage of the fact that the unknown current orientation at time $t + \delta t$ will be close to the orientation that was known at time $t$. However in order to be used in a certified manner, the Newton's method \emph{must} return a value that is \emph{unique} within its neighborhood: one way to ensure such a condition is the use of the \emph{Kantorovich unicity operator} \cite{Mer06}. In the sequel, the following notation will be used for a $(n\times n)$ matrix $\bm{M}\bydef\mat{M_{ij}}$ and a vector $\bm{x}$ of size $n$:
\begin{itemize}
	\setlength\itemsep{-0.15em}
	\item $\norm{\bm{x}}_\infty\bydef\max\limits_{i\in\itvd{1,n}}\abs{x_i}$ denotes the maximum norm (or $\infty$-norm) on $\mathbb{R}^n$. 
	\item $\norm{\bm{M}}_\infty\bydef\max\limits_{i\in\itvd{1,n}}\sum_{j=1}^n \abs{M_{ij}}$ denotes the row sum norm, an induced matrix norm on $\mathbb{R}^{n\times n}$.
\end{itemize}

The Newton-Kantorovich theorem \cite{Kan48,Tap71,DemMar73} states the Kantorovich test. Its aim is to investigate the existence and uniqueness of the root of $\bm{f}(\bm{x})=\bm{0}$ in a certain region. We will use the version formulated in \cite{DemMar73}.

\begin{theorem}[Newton-Kantorovich]
	Let $\bm{f}:\mathcal{D}\subseteq\mathbb{R}^n \to\mathbb{R}^n$ a function of class $\mathcal{C}^2$. Let $\bm{x}_0$ be a point and $\overline{\bm{U}}\left(\bm{x}_0\right)$ its neighborhood defined by $\overline{\bm{U}}\left(\bm{x}_0\right)\bydef\left\{\bm{x}\in \mathcal{D}\;\text{s.t.}\;\norm{\bm{x}-\bm{x}_0}_\infty\leq 2B_0\right\}$. Let $\bm{J}_0\bydef\bm{J}\left(\bm{x}_0\right)=\left.\partial\bm{f}/\partial\bm{x}\right|_{\bm{x}=\bm{x}_0}$ be an invertible jacobian matrix. If there exists three real constants $A_0$, $B_0$ and $C$ such that:
	\begin{enumerate}[label=\emph{(\roman*)}]
		\item $\norm{\bm{J}_0^{-1}}_\infty\leq A_0$
		\item $\norm{\bm{J}_0^{-1}\,\bm{f}\left(\bm{x}_0\right)}_\infty\leq B_0$
		\item $\forall\,i\in\itvd{1,n}, \forall\,j\in\itvd{1,n}$ and $\bm{x}\in\overline{\bm{U}}\left(\bm{x}_0\right),\;\displaystyle\sum_{k=1}^n \abs{\pdv[2]{f_i(\bm{x})}{x_j\partial x_k}}\leq C$
		\item $2nA_0B_0C\leq 1$
	\end{enumerate}
	then there is a unique solution of $\bm{f}(\bm{x})=\bm{0}$ in $\overline{\bm{U}}\left(\bm{x}_0\right)$ and the (real) Newton iterative scheme $\bm{x}_{k+1} = \bm{x}_{k}-\bm{J}^{-1}\left(\bm{x}_{k}\right)\,\bm{f}\left(\bm{x}_{k}\right)$ with the initial estimate $\bm{x}_0$ quadratically converges towards this unique solution. 
\end{theorem}

\begin{remark}
	A successful Kantorovich test is a \emph{sufficient} condition to certify the absence of any numerical instabilities (including singularities).
\end{remark}

If the Kantorovich test is valid, it provides a lower bound on the radius of the convergence domain towards the unique and guaranteed solution for Newton schemes. Hence, its pairing with a classical Newton scheme is in the heart of the certification of our SPM. In the case of the FGM certification, we have $\bm{x}\equiv\bm{o}$ and $\bm{J}\bydef\pdv{\bm{f}}{\bm{x}}\equiv\pdv{\bm{S}}{\bm{o}}$. The path tracking is initialized with the SPM's equilibrium configuration selected in the joint space analysis (see \eqref{eq:eq_config}) so that the initial orientation estimate $\bm{o}_0=\mathrm{FGM}{\left(\bm{j}_0\right)}$ is perfectly known. Then, for a small displacement from $\bm{j}_0$ to $\bm{j}_1\in\mathcal{Q}^\star$, we apply the Kantorovich test to the new coordinates. If valid, the above-mentionned test ensures the existence of an assembly mode $\bm{o}_1=\mathrm{FGM}{\left(\bm{j}_1\right)}$ and its uniqueness in a certain region that includes the convergence domain of the Newton scheme. Otherwise, the test is reapplied to coordinates that are closer to the last valid one. Finally, the same thing goes on for any displacement from $\bm{j}_k\in\mathcal{Q}^\star$ to $\bm{j}_{k+1}\in\mathcal{Q}^\star$.

%
	
	\subsection{Implementation of the Kantorovich test}

The path tracking strategy can be viewed as a semi-numerical approach to certify the robot (more precisely its FGM in our case). It is worth recalling that we manipulate a polynomial system $\bm{S}$ with integer coefficients. Morever, these coefficients are expressed with classical intervals to take into account the uncertainties.

\begin{definition}[Interval]
	An \emph{interval} $[x]$ is defined as the set of real numbers $x$ such that $\underline{x}\leq x\leq\overline{x}$. This interval has a \emph{width} $\wid([x])\bydef\overline{x}-\underline{x}$ and a \emph{midpoint} $\mpt([x])\bydef\left(\overline{x}+\underline{x}\right)/2$. The \emph{mignitude} (\textit{resp}. \emph{magnitude}) of $[x]$ is given by $\min{\left( \abs{\underline{x}},\abs{\overline{x}} \right)}$ (\textit{resp}. $\max{\left( \abs{\underline{x}},\abs{\overline{x}} \right)}$).
\end{definition}

\begin{figure}[htbp]
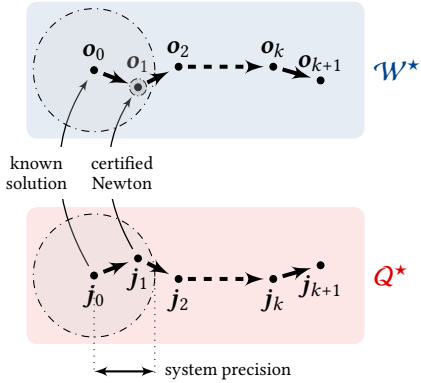

	\centering
	\includestandalone[]{TikZ/utilisation_kanto_interval}
	\caption{Implementation of the path tracking in orientation}
	\label{fig:impl_pt}
\end{figure}

In our implementation, the intervals are defined by the binary \emph{system precision} $\sigma$ such that $\wid{\left(\left[x\right]\right)}=1/2^{\sigma}$ and will be denoted as $\left[x\right]_{\sigma}$ in the sequel. The constants $A_0$, $B_0$ and $C$ are computed in a certified manner using \emph{multiple-precision arithmetic}: $A_0$ exclusively depends on the parallel Jacobian evaluated at the current coordinates, $B_0$ defines the neighborhood $\overline{\bm{U}}$ that will be used to determine $C$. The choice of multiple-precision (rather than double or floatting) arithmetic allows to perform any calculation on numbers whose digits of precision are limited only by the availiable memory of the device. Furthermore, the estimated orientations $\bm{o}_{k+1}$ returned by the certified Newton scheme are also computed using classical intervals. Figure \ref{fig:impl_pt} summarizes the way the path tracking in orientation is implemented.

Starting with $\bm{j}_0 = \mat{ \left[1\right]_9 & \left[1\right]_9 & \left[1\right]_9 }^{\mathsf{T}}$ and $\bm{o}_0=\mat{[0,0] & [0,0] & [0,0]}^{\mathsf{T}}$, we apply the path tracking in orientation to the neighborhood of $\bm{j}_0$ which is a ball $\mathcal{B}_0\subset\mathcal{Q}^\star$ containing $\bm{j}_0$ and whose size depends on the system precision. By taking into account all the possible values of $\bm{j}_1\in\mathcal{B}_0$, we are considering a family of infinite systems in $\mathcal{B}_0$. The success of the Kantorovich test certifies the computation of $\mathrm{FGM}\left(\bm{j}_1\right)=\bm{o}_1$, $\forall\,\bm{j}_1\in\mathcal{B}_0$. The ball $\mathcal{B}'_0\subset\mathcal{W}^\star$ containing all the solutions $\bm{o}_1$ is included in the convergence domain (provided by the Kantorovich test) that contains the initial estimate $\bm{o}_0$. Applying the Kantorovich test directly in the joint space may generate lots of computations as we deal with three variables $j_1$, $j_2$ and $j_3$. Part of the implementation strategy is also to reduce the computational cost by considering the geometrical simplifications of our mechanism. Indeed, by taking into account the invariance w.r.t. $o_3\in\mathcal{W}$, two coordinates ($o_1$ and $o_2$) are sufficient to detect any problematic configurations. Thus, the only values considered in the joint space are the ones coming from the computation of $\bm{j}=\mathrm{IGM}\left(o_1,o_2,o_3=0\right)$ A scanning of our workspace $\mathcal{W}^\star\left(o_3=0\right)$ proves that the Kantorovich test is valid everywhere, despite a poor binary system precision of $\sigma=9$ bits and a displacement step of $\frac{1}{100}$ in $\mathcal{W}^\star$. This certifies the FGM of our robot given our applications.


\section{Conclusion and further works}

In this paper, we have presented two approaches to certify the symetrical spherical parallel manipulator with coaxial input shafts. The first approach is the symbolic one involving the computation of the discriminant variety w.r.t. the projection onto the parameter space for the inverse model. The explicit expressions obtained allowed us to clearly determine a set in the orientation space free of Type-1 singularities (and other numerical instabilities) that includes our prescribed workspace. This strategy could not (yet) be applied to the forward model as its coefficients and degrees are much bigger compared to the inverse ones. Therefore, we used a semi-numerical approach involving the Kantorovich unicity operator and a classical Newton scheme to certify the forward model with a successful path tracking in orientation. The numerical computations were done here considering uncertainties on the fabrication parameters that are translated into uncertainties on the coefficients of our system with interval and multiple precision arithmetic. These certification tools and strategies could naturally be extended to any other parallel robot. This work was also the opportunity to apprehend the behavior of our mechanism in terms of motion with the computation of the joint stops. Further works will use the basic concepts of this article to extend the study to other spherical parallel manipulators with different conception parameters.

\begin{acks}
We thank Jean-Pierre Merlet and Clément Gosselin for the fruitful discussions and their feedbacks regarding the parallel mechanism of interest. We are also grateful to people we interacted with for this article.
\end{acks}

\bibliographystyle{apalike}
\bibliography{IEEEabrv,ref}

\appendix

\section{Determination of the SPM design using the GCI approach}\label{sec:a1}

The \emph{global conditioning index} $\mathrm{GCI}$ is \emph{numerically} computed using the following kinematic criteria
\begin{equation}
	\mathrm{GCI} \bydef \dfrac{\iint_{\mathcal{W}^\star} \zeta{\left(\bm{J}\right)} \dd{\chi_1}\dd{\chi_2}}{\iint_{\mathcal{W}^\star} \dd{\chi_1}\dd{\chi_2} }\hspace*{3em}\text{where}\quad\zeta(\bm{J}) \bydef \dfrac{1}{\kappa{\left(\bm{J}\right)}} = \dfrac{1}{\norm{\bm{J}}\,\norm{\bm{J}^{-1}}}
\end{equation}
where $\bm{J}$ being the SPM's \emph{Jacobian matrix} depending on $\bm{\chi}$, $\bm{\theta}$ and conception parameters of Tab. \ref{tab:parameters}, $\norm{\bm{J}}$ its Frobenius norm defined by $\norm{\bm{J}} \bydef \tr^{1/2}{\left( \bm{J}^{\mathsf{T}}\,\frac{1}{n_a}\id_{n_a}\,\bm{J} \right)}$ and $0\leq\zeta{\left(\bm{J}\right)}\leq1$ its conditioning index. With such a method and $80\times80$ points, we get $\mathrm{GCI}=0.93$, $\zeta_{\min}=0.8902$ and $\zeta_{\max}=0.9487$.

\section{conditioning index of the Jacobian matrices}\label{sec:a3}

\begin{figure}[htbp]
	\centering
	\begin{subfigure}{0.49\textwidth}
        \centering
        \includegraphics[scale=0.475]{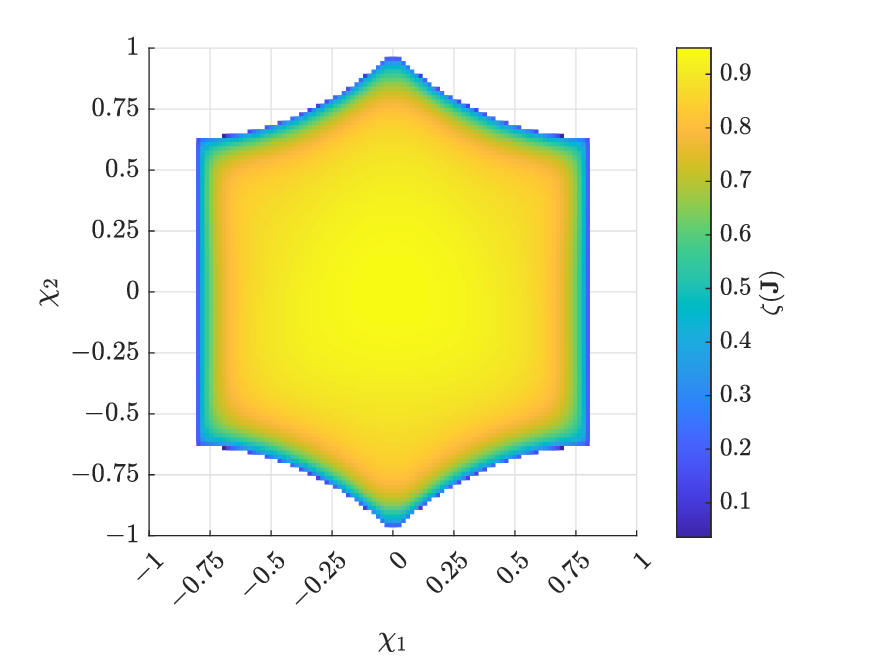}
        \caption{$\zeta{\left(\bm{J}\right)}$}
        \label{fig:ci_global}
    \end{subfigure}
    \begin{subfigure}{0.49\textwidth}
        \centering
        \includegraphics[scale=0.475]{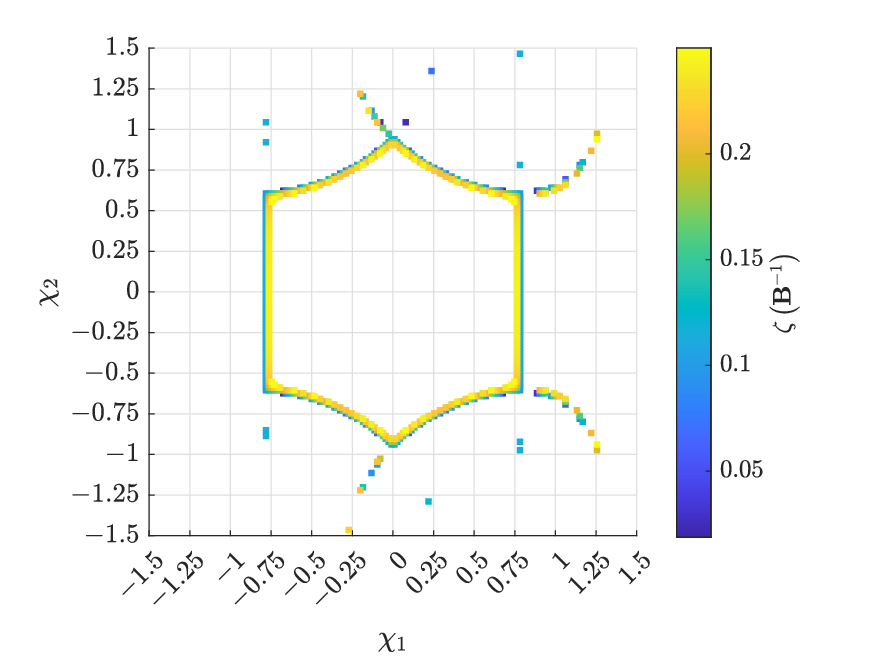}
        \caption{$\zeta{\left(\bm{B}\right)}<0.25$ (Type-1 singularities)}
        \label{fig:ci_type1}
    \end{subfigure}
	
	\caption{Conditioning index of the Jacobian matrices of the SPM}
	\label{fig:ci_overview}
\end{figure}

\section{Geometric model of the SPM in its polynomial form}\label{sec:a2}

\begin{equation}
	\bm{S}\bydef\left\{ \begin{aligned}
		-\sqrt{2}\, j_{1}^{2} o_{1}^{2} o_{3}^{2}-2 \sqrt{2}\, j_{1}^{2} o_{1} o_{3}^{2}+\sqrt{2}\, j_{1}^{2} o_{1}^{2}+\sqrt{2}\, j_{1}^{2} o_{3}^{2}+4 \sqrt{2}\, j_{1} o_{1}^{2} o_{3}+\sqrt{2}\, o_{1}^{2} o_{3}^{2}\\
		-2 \sqrt{2}\, j_{1}^{2} o_{1}-2 \sqrt{2}\, o_{1} o_{3}^{2}-\sqrt{2}\, j_{1}^{2}-4 \sqrt{2}\, j_{1} o_{3}-\sqrt{2}\, o_{1}^{2}-\sqrt{2}\, o_{3}^{2}-2 \sqrt{2}\, o_{1}+\sqrt{2} &= 0\\[2ex]
		\sqrt{2}\, o_{1}^{2}-2 \sqrt{2}\, o_{3}^{2}+2 \sqrt{2}\, o_{1}-2 \sqrt{2}\, j_{2}^{2}-\sqrt{2}\, o_{2}^{2}+2 \sqrt{2}\, \sqrt{3}\, j_{2}^{2} o_{1}^{2} o_{2}-2 \sqrt{2}\, \sqrt{3}\, j_{2}^{2} o_{1}^{2} o_{3}\\
		+2 \sqrt{2}\, \sqrt{3}\, j_{2}^{2} o_{2}^{2} o_{3}-2 \sqrt{2}\, \sqrt{3}\, j_{2}^{2} o_{2} o_{3}^{2}\\
		-2 \sqrt{2}\, \sqrt{3}\, j_{2} o_{1}^{2} o_{3}^{2}+2 \sqrt{2}\, \sqrt{3}\, j_{2} o_{2}^{2} o_{3}^{2}+2 \sqrt{2}\, \sqrt{3}\, o_{1}^{2} o_{2} o_{3}^{2}-2 \sqrt{2}\, \sqrt{3}\, j_{2}^{2} o_{1} o_{2}\\
		-2 \sqrt{2}\, \sqrt{3}\, o_{1} o_{2} o_{3}^{2}-2 \sqrt{2}\, j_{2}^{2} o_{1}^{2} o_{2}^{2} o_{3}^{2}+12 \sqrt{2}\, j_{2}^{2} o_{1} o_{2} o_{3}+8 \sqrt{2}\, j_{2} o_{1}^{2} o_{2}^{2} o_{3}\\
		+12 \sqrt{2}\, j_{2} o_{1} o_{2} o_{3}^{2}+2 \sqrt{2}\, j_{2}^{2} o_{1} o_{2}^{2} o_{3}^{2}-\sqrt{2}\, o_{1}^{2} o_{3}^{2}+2 \sqrt{2}\, o_{1} o_{3}^{2}-\sqrt{2}\, j_{2}^{2} o_{1}^{2}+\sqrt{2}\, j_{2}^{2} o_{2}^{2}\\
		+2 \sqrt{2}\, j_{2}^{2} o_{3}^{2}-8 \sqrt{2}\, j_{2} o_{3}+2 \sqrt{2}\, j_{2}^{2} o_{1}+2 \sqrt{2}\, o_{1} o_{2}^{2}-2 \sqrt{2}\, \sqrt{3}\, o_{2}-2 \sqrt{2}\, o_{1}^{2} o_{2}^{2}+\sqrt{2}\, o_{2}^{2} o_{3}^{2}\\
		+2 \sqrt{2}\, \sqrt{3}\, j_{2}^{2} o_{1}^{2} o_{2} o_{3}^{2}+2 \sqrt{2}\, \sqrt{3}\, j_{2}^{2} o_{1} o_{2} o_{3}^{2}-8 \sqrt{2}\, \sqrt{3}\, j_{2} o_{1} o_{2} o_{3}-12 \sqrt{2}\, o_{3} o_{1} o_{2}\\
		+2 \sqrt{2}\, j_{2}^{2} o_{1}^{2} o_{2}^{2}+\sqrt{2}\, j_{2}^{2} o_{1}^{2} o_{3}^{2}-\sqrt{2}\, j_{2}^{2} o_{2}^{2} o_{3}^{2}-4 \sqrt{2}\, j_{2} o_{1}^{2} o_{3}+4 \sqrt{2}\, j_{2} o_{2}^{2} o_{3}\\
		-12 \sqrt{2}\, j_{2} o_{1} o_{2}+2 \sqrt{2}\, j_{2}^{2} o_{1} o_{2}^{2}+2 \sqrt{2}\, j_{2}^{2} o_{1} o_{3}^{2}+2 \sqrt{2}\, o_{1} o_{2}^{2} o_{3}^{2}-2 \sqrt{2}\, \sqrt{3}\, j_{2}^{2} o_{2}\\
		+2 \sqrt{2}\, \sqrt{3}\, j_{2} o_{1}^{2}-2 \sqrt{2}\, \sqrt{3}\, j_{2} o_{2}^{2}+2 \sqrt{2}\, \sqrt{3}\, o_{1}^{2} o_{2}+2 \sqrt{2}\, \sqrt{3}\, o_{1}^{2} o_{3}-2 \sqrt{2}\, \sqrt{3}\, o_{2}^{2} o_{3}\\
		-2 \sqrt{2}\, \sqrt{3}\, o_{2} o_{3}^{2}+2 \sqrt{2}\, \sqrt{3}\, o_{1} o_{2}+2 \sqrt{2}\, o_{1}^{2} o_{2}^{2} o_{3}^{2}+2 \sqrt{2} &= 0\\[2ex]
		\sqrt{2}\, o_{1}^{2}-2 \sqrt{2}\, o_{3}^{2}+2 \sqrt{2}\, o_{1}-\sqrt{2}\, o_{2}^{2}-2 \sqrt{2}\, j_{3}^{2}+8 \sqrt{2}\, j_{3} o_{1}^{2} o_{2}^{2} o_{3}+12 \sqrt{2}\, j_{3} o_{1} o_{2} o_{3}^{2}\\
		+2 \sqrt{2}\, j_{3}^{2} o_{1} o_{2}^{2} o_{3}^{2}-2 \sqrt{2}\, \sqrt{3}\, j_{3}^{2} o_{1}^{2} o_{2}+2 \sqrt{2}\, \sqrt{3}\, j_{3}^{2} o_{1}^{2} o_{3}-2 \sqrt{2}\, \sqrt{3}\, j_{3}^{2} o_{2}^{2} o_{3}\\
		-2 \sqrt{2}\, \sqrt{3}\, o_{1}^{2} o_{2} o_{3}^{2}+2 \sqrt{2}\, \sqrt{3}\, o_{1} o_{2} o_{3}^{2}+2 \sqrt{2}\, \sqrt{3}\, j_{3}^{2} o_{2} o_{3}^{2}+2 \sqrt{2}\, \sqrt{3}\, j_{3} o_{1}^{2} o_{3}^{2}\\
		-2 \sqrt{2}\, \sqrt{3}\, j_{3} o_{2}^{2} o_{3}^{2}+2 \sqrt{2}\, \sqrt{3}\, j_{3}^{2} o_{1} o_{2}-2 \sqrt{2}\, j_{3}^{2} o_{1}^{2} o_{2}^{2} o_{3}^{2}+12 \sqrt{2}\, j_{3}^{2} o_{1} o_{2} o_{3}\\
		-\sqrt{2}\, j_{3}^{2} o_{1}^{2}+\sqrt{2}\, j_{3}^{2} o_{2}^{2}+2 \sqrt{2}\, j_{3}^{2} o_{3}^{2}-8 \sqrt{2}\, j_{3} o_{3}+2 \sqrt{2}\, j_{3}^{2} o_{1}-\sqrt{2}\, o_{1}^{2} o_{3}^{2}+2 \sqrt{2}\, o_{1} o_{3}^{2}\\
		+2 \sqrt{2}\, o_{1} o_{2}^{2}+2 \sqrt{2}\, \sqrt{3}\, o_{2}-2 \sqrt{2}\, o_{1}^{2} o_{2}^{2}+\sqrt{2}\, o_{2}^{2} o_{3}^{2}-2 \sqrt{2}\, \sqrt{3}\, j_{3}^{2} o_{1}^{2} o_{2} o_{3}^{2}\\
		-2 \sqrt{2}\, \sqrt{3}\, j_{3}^{2} o_{1} o_{2} o_{3}^{2}+8 \sqrt{2}\, \sqrt{3}\, j_{3} o_{1} o_{2} o_{3}-12 \sqrt{2}\, o_{3} o_{1} o_{2}+2 \sqrt{2}\, o_{1} o_{2}^{2} o_{3}^{2}\\
		-2 \sqrt{2}\, \sqrt{3}\, o_{1}^{2} o_{2}-2 \sqrt{2}\, \sqrt{3}\, o_{1}^{2} o_{3}+2 \sqrt{2}\, \sqrt{3}\, o_{2}^{2} o_{3}+2 \sqrt{2}\, \sqrt{3}\, o_{2} o_{3}^{2}-2 \sqrt{2}\, \sqrt{3}\, o_{1} o_{2}\\
		+2 \sqrt{2}\, o_{1}^{2} o_{2}^{2} o_{3}^{2}-4 \sqrt{2}\, j_{3} o_{1}^{2} o_{3}+4 \sqrt{2}\, j_{3} o_{2}^{2} o_{3}-12 \sqrt{2}\, j_{3} o_{1} o_{2}+2 \sqrt{2}\, j_{3}^{2} o_{1} o_{2}^{2}\\
		+2 \sqrt{2}\, j_{3}^{2} o_{1} o_{3}^{2}+2 \sqrt{2}\, \sqrt{3}\, j_{3}^{2} o_{2}-2 \sqrt{2}\, \sqrt{3}\, j_{3} o_{1}^{2}+2 \sqrt{2}\, \sqrt{3}\, j_{3} o_{2}^{2}+2 \sqrt{2}\, j_{3}^{2} o_{1}^{2} o_{2}^{2}\\
		+\sqrt{2}\, j_{3}^{2} o_{1}^{2} o_{3}^{2}-\sqrt{2}\, j_{3}^{2} o_{2}^{2} o_{3}^{2}+2 \sqrt{2} &= 0
	\end{aligned}\right.
\end{equation}

\end{document}

%% file: packages/math.tex
\usepackage{bigints} 

\usepackage{mathtools}

\newcommand{\bydef}{\triangleq}

\usepackage{xstring}
\noexpandarg 

	\DeclareRobustCommand{\dder}{
		\mathop{}\mathopen{}\mathrm{d}
	}

	\newcommand{\dd}[2][0]{
		\IfStrEq{#1}{0}{
			\dder #2
		}{
			\IfBeginWith{#2}{f}{
				\dder^{#1} \! #2
			}{
				\dder^{#1}  #2
			}
		}
	}

	\newcommand{\dv}[3][0]{
		\IfStrEq{#1}{0}{
			\ensuremath{\frac{\dd{#2}}{\dd{#3}}}
		}{
			\ensuremath{\frac{\dd[#1]{#2}}{\dd{#3}^{#1}}}
		}
	}

	\makeatletter
		\let\original@partial\partial
		\renewcommand{\partial}{
			\original@partial\mathopen{}
		}
	\makeatother

	\newcommand{\pp}[2][0]{
		\IfStrEq{#1}{0}{
			\partial #2
		}{
			\IfBeginWith{#2}{f}{
				\partial^{#1} \! #2
			}{
				\partial^{#1} #2
			}
		}
	}

	\newcommand\pdv[3][0]{%
		\def\deno{\StrSubstitute{ #3}{ }\partial}
		\def\denoOK{\StrSubstitute{\deno}{\partial f}{\partial \! f}}
		\IfStrEq{#1}{0}{
			\ensuremath{\frac{\pp{#2}}{\denoOK}}
		}{
			\ensuremath{\frac{\pp^{#1}{#2}}{\denoOK}}
		}
	}

\DeclareMathOperator{\res}{Res}

\DeclareMathOperator{\discrim}{discrim}
\DeclareMathOperator{\proj}{proj}

\DeclareMathOperator{\LC}{LC}

\DeclareMathOperator{\tr}{Tr}
\DeclareMathOperator{\id}{\mathbf{1}}

\usepackage{stmaryrd} 
\SetSymbolFont{stmry}{bold}{U}{stmry}{m}{n}
\usepackage{bm}

\newcommand{\rep}[2]{\prescript{#2}{}{#1}}

\DeclareMathOperator{\wid}{width}
\DeclareMathOperator{\mpt}{mid}

\newcommand\norm[1]{\left\lVert#1\right\rVert}

\newcommand\abs[1]{\left|#1\right|}

\newcommand\mat[1]{\left[\;\begin{matrix}#1\end{matrix}\;\right]}
\newcommand\bra[1]{\left\langle#1\right\rangle}
\newcommand\itvd[1]{\left\llbracket#1\right\rrbracket}

\newcommand{\1}{\text{\usefont{U}{bbold}{m}{n}1}}
\MakeRobust{\1}

%% file: packages/add_ons.tex
\usepackage{lipsum}

\usepackage{multirow}
\usepackage{hhline}

\usepackage{subcaption}

\usepackage{framed}

\usepackage{qrcode}

\usepackage{empheq}
\usepackage{fancybox,xcolor}

\definecolor{shadowcolor}{RGB}{0, 0, 102}

\usepackage[french]{algorithm2e}
\SetKwInput{KwRes}{R\'esultat}%
\SetKwIF{Si}{SinonSi}{Sinon}{si}{alors}{sinon si}{sinon}{fin si}%
\SetKwFor{Tq}{tant que}{faire}{fin tq}%
\SetKwRepeat{FTq}{faire}{tant que}
\usepackage{algorithmic}

\usepackage{siunitx}

\usepackage{enumitem}

\usepackage{standalone}

\AtEndDocument{\label{lastpage}}

\usepackage{multicol}
\usepackage{pdfpages}

\usepackage[]{nomencl}
\makenomenclature
\usepackage{etoolbox}
\renewcommand\nomgroup[1]{%
  \item[\bfseries
  \ifstrequal{#1}{A}{Sets}{%
  \ifstrequal{#1}{B}{Mathematical objects}{%
  \ifstrequal{#1}{C}{Robotics \& Mechanics}{%
  \ifstrequal{#1}{D}{Interval analysis}{}}}}%
]}

\usepackage[percent]{overpic} 

\newtheorem{theorem}{Theorem}[]
\newtheorem{definition}{Definition}[]

\newtheorem{remark}{Remark}[]

%% file: packages/draw.tex
\usepackage{xcolor}
\definecolor{main}{RGB}{0,0,0}
\definecolor{dgreen}{RGB}{0,111,0}
\definecolor{dblue}{RGB}{0,0,160}
\definecolor{dred}{RGB}{223,0,0}
\definecolor{safran}{RGB}{20,119,198}
\definecolor{bleu_identitaire}{RGB}{1,68,145}
\definecolor{gris_identitaire}{RGB}{82,86,104}

\usepackage{pgf,pgfplots,tikz}
\usetikzlibrary{calc,arrows}
\pgfplotsset{width=10cm,compat=1.9}
\usepackage{tikz-cd}
\usepackage{tikz-3dplot}
\usepackage{graphicx}

\usepackage[oldvoltagedirection]{circuitikz}
\usetikzlibrary{babel}

\usepackage{schemabloc}
\usepackage{bodegraph}

\usetikzlibrary{decorations.pathmorphing,patterns}
\usetikzlibrary{automata,positioning}

\tikzstyle{block} = [draw, fill=white, rectangle, 
        minimum height=3em, minimum width=3em]
\tikzstyle{mblock} = [draw, fill=white, rectangle, 
        minimum height=1.75em, minimum width=3em]       
\tikzstyle{tblock} = [draw, fill=white, rectangle,       
        minimum height=5em, minimum width=3em]          
\tikzstyle{bblock} = [draw, fill=white, rectangle,       
        minimum height=10em, minimum width=3em]          
\tikzstyle{sblock} = [draw, fill=white, rectangle, 
        minimum height=1.5em, minimum width=3em]
\tikzstyle{sum} = [draw, fill=white, circle, node distance=1cm]
\tikzstyle{input} = [coordinate]
\tikzstyle{output} = [coordinate]
\tikzstyle{pinstyle} = [pin edge={to-,thin,black}]

\tikzstyle{decision} = [diamond, draw, fill=blue!10, 
    text width=7.5em, text badly centered, node distance=4cm, inner sep=0pt]
\tikzstyle{etape} = [rectangle, draw, fill=blue!10, 
    text centered, rounded corners, minimum height=4em]
\tikzstyle{line} = [draw, -latex']
\tikzstyle{cloud} = [draw, ellipse,fill=red!10, node distance=3cm,
    minimum height=2em]

\newcommand{\nvar}[2]{%
    \newlength{#1}
    \setlength{#1}{#2}
}

\nvar{\dg}{0.3cm}

\nvar{\ddx}{1.15cm} 

\pgfdeclarelayer{background}
\pgfdeclarelayer{foreground}
\pgfsetlayers{background,main,foreground}

\usepackage{stackengine}
\usepackage{scalerel}

\pgfmathsetmacro{\xdeg}{30}
\pgfmathsetmacro{\xx}{cos(\xdeg)}
\pgfmathsetmacro{\xy}{sin(\xdeg)}

\pgfmathsetmacro{\ydeg}{120}
\pgfmathsetmacro{\yx}{cos(\ydeg)}
\pgfmathsetmacro{\yy}{sin(\ydeg)}

\pgfmathsetmacro{\zdeg}{80}
\pgfmathsetmacro{\zx}{cos(\zdeg)}
\pgfmathsetmacro{\zy}{sin(\zdeg)}

\newcommand{\tdcyl}[5]{
    \path (1,0,0);
    \pgfgetlastxy{\cylxx}{\cylxy}
    \path (0,1,0);
    \pgfgetlastxy{\cylyx}{\cylyy}
    \path (0,0,1);
    \pgfgetlastxy{\cylzx}{\cylzy}
    \pgfmathsetmacro{\cylt}{(\cylzy * \cylyx - \cylzx * \cylyy)/ (\cylzy * \cylxx - \cylzx * \cylxy)}
    \pgfmathsetmacro{\ang}{atan(\cylt)}
    \pgfmathsetmacro{\ct}{1/sqrt(1 + (\cylt)^2)}
    \pgfmathsetmacro{\st}{\cylt * \ct}
    \filldraw[fill=white] (#4*\ct+#1,#4*\st+#2,#3) -- ++(0,0,#5) arc[start angle=\ang,delta angle=-180,radius=#4] -- ++(0,0,-#5) arc[start angle=\ang+180,delta angle=180,radius=#4];
    \filldraw[fill=white] (#1,#2,#3+#5) circle[radius=#4];
    \filldraw[fill=white] (#1,#2,#3) circle[radius=#4];
}
\usetikzlibrary{spy}

%% file: TikZ/robot_parallele_structure.tex
	\begin{tikzpicture}[>=stealth,every text node part/.style={align=center}]
		\node[line width=0.05em, ellipse, draw = black!40, minimum width = 12em, minimum height = 4em] (inf) at (0,0) {};
		
		\draw[draw=black,fill=black!0] (inf.west)node(l11){}node[left]{\tiny $A_{11}$} circle (0.25em);
		\draw[draw=black,fill=black!0] (inf.-140)node(l21){}node[left]{\tiny $A_{21}$} circle (0.25em);
		\draw[draw=black,fill=black!0] (inf.-40)node(l31){}node[right]{\tiny $A_{k1}$} circle (0.25em);
		\draw[draw=black,fill=black!0] (inf.east)node(l41){}node[right]{\tiny $A_{n1}$} circle (0.25em);
		
		\foreach \jj in {1,2}{
		\foreach \ii/\kk/\ll[count=\ci from 2,count=\di from 1] in{-1.05/40/2,-1.5/0/3,-1.5/0/(m_\jj-2),-1.05/0/(m_\jj-1),0/0/m_\jj}{
			\draw[draw=black,fill=black!\kk] ($(l\jj1)+(\ii em,1.5*\di em)$) node(l\jj\ci){}node[left]{\tiny $A_{\jj\ll}$} circle (0.25em);
		}}
		\foreach \jj/\mm in {3/k,4/n}{
		\foreach \ii/\kk/\ll[count=\ci from 2,count=\di from 1] in{1.05/40/2,1.5/0/3,1.5/0/(m_\mm-2),1.05/0/(m_\mm-1),0/0/m_\mm}{
			\draw[draw=black,fill=black!\kk] ($(l\jj1)+(\ii em,1.5*\di em)$) node(l\jj\ci){}node[right]{\tiny $A_{\mm\ll}$} circle (0.25em);
		}}
		
		
		\begin{pgfonlayer}{background}
			\draw[line width=0.2em] (inf.west)--($(inf.west)+(-1.05 em,1.5*1 em)$)--($(inf.west)+(-1.5 em,1.5*2 em)$) ($(inf.west)+(-1.5 em,1.5*3 em)$)--($(inf.west)+(-1.05 em,1.5*4 em)$)--($(inf.west)+(0 em,1.5*5 em)$);
			\draw[line width=0.1em,dotted]($(inf.west)+(-1.5 em,1.5*2 em)$)--($(inf.west)+(-1.5 em,1.5*3 em)$);		
			
			\draw[line width=0.2em] (inf.-140)--($(inf.-140)+(-1.05 em,1.5*1 em)$)--($(inf.-140)+(-1.5 em,1.5*2 em)$) ($(inf.-140)+(-1.5 em,1.5*3 em)$)--($(inf.-140)+(-1.05 em,1.5*4 em)$)--($(inf.-140)+(0 em,1.5*5 em)$);
			\draw[line width=0.1em,dotted]($(inf.-140)+(-1.5 em,1.5*2 em)$)--($(inf.-140)+(-1.5 em,1.5*3 em)$);
			
			\draw[line width=0.2em] (inf.-40)--($(inf.-40)+(1.05 em,1.5*1 em)$)--($(inf.-40)+(1.5 em,1.5*2 em)$) ($(inf.-40)+(1.5 em,1.5*3 em)$)--($(inf.-40)+(1.05 em,1.5*4 em)$)--($(inf.-40)+(0 em,1.5*5 em)$);
			\draw[line width=1em,draw=dgreen!25,draw opacity=0.5,text opacity=1] (inf.-40)node[below,dgreen]{$k$\textsuperscript{th} kinematic chain\\(a robot leg)}--($(inf.-40)+(1.05 em,1.5*1 em)$)--($(inf.-40)+(1.5 em,1.5*2 em)$)--($(inf.-40)+(1.5 em,1.5*3 em)$)--($(inf.-40)+(1.05 em,1.5*4 em)$)--($(inf.-40)+(0 em,1.5*5 em)$);
			\draw[line width=0.1em,dotted]($(inf.-40)+(1.5 em,1.5*2 em)$)--($(inf.-40)+(1.5 em,1.5*3 em)$);
			
			\draw[line width=0.2em] (inf.east)--($(inf.east)+(1.05 em,1.5*1 em)$)--node[midway](body){}($(inf.east)+(1.5 em,1.5*2 em)$) ($(inf.east)+(1.5 em,1.5*3 em)$)--($(inf.east)+(1.05 em,1.5*4 em)$)--($(inf.east)+(0 em,1.5*5 em)$);
			\draw[line width=0.1em,dotted]($(inf.east)+(1.5 em,1.5*2 em)$)--($(inf.east)+(1.5 em,1.5*3 em)$);
		\end{pgfonlayer}
		
		\begin{pgfonlayer}{background}
			\node[rotate=0,line width=0.05em, ellipse, draw = black!40, minimum width = 12em, minimum height = 4em] (sup) at (0,7.5em) {};
		\end{pgfonlayer}
		
		\draw[->,thick,blue] (inf.center)--($(inf.center)+(-0.37em,-0.7em)$)node[below=-0.25em]{\scriptsize $\bm{y}_0$};
		\draw[->,thick,blue] (inf.center)--($(inf.center)+(1em,0 em)$)node[right=-0.25em]{\scriptsize $\bm{x}_0$};
		\draw[->,thick,blue] (inf.center)node[left]{\tiny$O$}--($(inf.center)+(0 em,1 em)$)node[above=-0.25em]{\scriptsize $\bm{z}_0$};
		
		\draw[->,thick,red] (sup.center)--($(sup.center)+(-0.37em,-0.7em)$)node[below=-0.25em]{\scriptsize $\bm{y}_p$};
		\draw[->,thick,red] (sup.center)--($(sup.center)+(1em,0 em)$)node[right=-0.25em]{\scriptsize $\bm{x}_p$};
		\draw[->,thick,red] (sup.center)node[left]{\tiny$P$}--($(sup.center)+(0 em,1 em)$)node[above=-0.25em]{\scriptsize $\bm{z}_p$};
		
		\draw[<-,thick] (sup.50)to[in=180,out=90]($(sup.50)+(2.5em,1.25em)$)node[right]{Moving platform};
		\draw[<-,thick] (inf.-10)to[in=180,out=-90]($(inf.-10)+(2.5em,-1.25em)$)node[right]{Base};
		\draw[<-,thick,blue] (l11)to[in=0,out=-120]($(l11)+(-3.5em,-1.25em)$)node[left]{Passive joint};
		\draw[<-,thick,red] (l12)to[in=0,out=-120]($(l12)+(-3.5em,-1.25em)$)node[left]{Active joint};
		\draw[<-,thick,safran] ($(body)+(0.1em,0em)$)--($(body)+(3.5em,0em)$)node[right]{Body};
	\end{tikzpicture}

%% file: TikZ/poignet_agile_sc_rep_with_etas.tex
    \tdplotsetmaincoords{70}{60+90}
    
    \begin{tikzpicture}[scale=3,tdplot_main_coords,>=stealth,every text node part/.style={align=center}]
    
        \pgfmathsetmacro{\cx}{0.3}
        \pgfmathsetmacro{\cy}{0.15}
        \pgfmathsetmacro{\cz}{0.15}
    
        \draw[line width=0.175em,fill=black!10] (0,1.5,-1.5)-- (1.5*sin(120,1.5*cos(120,-1.5)node[left=0.25em]{$B_i$}--(1.5*sin(240,1.5*cos(240,-1.5)--cycle;
        
        \draw[thick] (0,1.5,-1.5)--(0,0,1.5-1.5);
        \draw[thick] (1.5*sin(120,1.5*cos(120,-1.5)--(0,0,1.5-1.5);
        \draw[thick] (1.5*sin(240,1.5*cos(240,-1.5)--(0,0,1.5-1.5)node[above,right=1em]{$O$};
        
        \foreach \i in{0,120,240}{
        	\draw[dashed] (0,0,-1.5)--(1.5*sin(\i,1.5*cos(\i,-1.5);
        }
        \node at (0,0,-1.5) (centerB) {};
        \node[below of=centerB,below=0.5em] (ob) {$O_B$};
        \draw[very thin,->]  (ob)to[out=100,in=-100](centerB);
    
        \draw[red, ->, fill opacity=0.45, draw opacity=0.45] (0,0,0-1.5) -- (1.75,0,0-1.5) node[anchor=east]{$\bm{x}_0$};
        \draw[red, ->, fill opacity=0.45, draw opacity=0.45] (0,0,0-1.5) -- (0,2,0-1.5) node[anchor=west]{$\bm{y}_0$};
        \draw[red, ->, fill opacity=0.45, draw opacity=0.45] (0,0,0-1.5) -- (0,0,2-1.5) node[anchor=south]{$\bm{z}_0$};
        
        \draw[dashed] (0,0,-1.5)--(0,0,0);
        
        \begin{scope}[rotate around z=240]
            \draw[thick,blue,->] (0,1.5,0-1.5)--(0,2,-0.5-1.5)node[anchor=north]{$\bm{z}_{1,i}\equiv\bm{u}_i$};
        	\draw[thick,blue!50,->] (0,1.5,0-1.5)--(0.65,1.5,-1.5)node[anchor=south]{$\bm{x}_{1,i}$};
        	\draw[thick,blue!50,->] (0,1.5,0-1.5)--(0,1.17500000000000,-1.82500000000000)node[anchor=north]{$\bm{y}_{1,i}$};
        \end{scope}
        
        \tdplotdefinepoints(0,0,0)(0,0,-1.5)(0,1.5,-1.5)
        \tdplotdrawpolytopearc[thick,dgreen]{1}{anchor=south}{\color{dgreen}$\beta_1$}
        
        \tdplotdefinepoints(0,0,0)(1.5*sin(120),1.5*cos(120),-1.5)(0,1.5,-1.5)
        \tdplotdrawpolytopearc[thick,dgreen]{0.5}{anchor=south}{\color{dgreen}$\gamma_1$}

        \begin{scope}[rotate around x=170,rotate around z=-180,rotate around y=-20]
            \draw[line width=0.175em,fill=black!10] (0,1.5,-1.5)-- (1.5*sin(120,1.5*cos(120,-1.5)--(1.5*sin(240,1.5*cos(240,-1.5)node[left]{$A_i$}--cycle;
        
            \draw[thick] (0,1.5,-1.5)--(0,0,1.5-1.5);
            \draw[thick] (1.5*sin(120,1.5*cos(120,-1.5)--(0,0,1.5-1.5);
            \draw[thick] (1.5*sin(240,1.5*cos(240,-1.5)--(0,0,1.5-1.5);
            
            \foreach \i in{0,120,240}{
            	\draw[dashed] (0,0,-1.5)--(1.5*sin(\i,1.5*cos(\i,-1.5);
            }
            \node at (0,0,-1.5) (centerA) {};
            \node[above of=centerA,above=1.5em] (oa) {$O_A$};
            \draw[very thin,->]  (oa)to[out=-105,in=105](centerA);
            
            \draw[dashed] (0,0,-1.5)--(0,0,0);
            \draw[thick, safran, ->] (0,0,-1.5)--(0,0,-2)node[above]{$\bm{n}$};
            
            \begin{scope}[rotate around z=120]
                \draw[thick,blue,->] (0,1.5,0-1.5)--(0,2,-0.5-1.5)node[anchor=south]{$\bm{z}_{3,i}\equiv\bm{v}_i$};
        		\draw[thick,blue!50,->] (0,1.5,0-1.5)--(0.65,1.5,-1.5)node[anchor=north]{$\bm{y}_{3,i}$};
        		\draw[thick,blue!50,->] (0,1.5,0-1.5)--(0,1.17500000000000,-1.82500000000000)node[anchor=south,right=0.25em]{$\bm{x}_{3,i}$};
            \end{scope}
            
            \begin{scope}[rotate around z=120,rotate around x=45]
                \tdcyl{0}{0}{-1.5}{0.05}{0.3}
                \node at (0-0.01,0-0.05,-1.5+0.3/2) (l3) {}; 
            \end{scope}
        \end{scope}
        
        \tdplotdefinepoints(0,0,0)(-.5130302150,1.721975497,1.127652601)(0.7076663070,-.4166905490,1.955908328)
        \tdplotdrawpolytopearc[thick,dgreen]{0.5}{anchor=south}{\color{dgreen}$\gamma_2$}
            
        \tdplotdefinepoints(0,0,0)(-.5130302150,0.2447638668,1.388124868)(0.7076663070,-.4166905490,1.955908328)
        \tdplotdrawpolytopearc[thick,dgreen]{0.75}{anchor=south}{\color{dgreen}$\beta_2$}
        
        \begin{scope}[rotate around z=240,rotate around x=45]
            \tdcyl{0}{0}{-1.5}{0.05}{0.3}
            \node at (0-0.01,0-0.05,-1.5+0.3/2) (l1) {}; 
        \end{scope}
        
        \begin{scope}[rotate around y=-60,rotate around z=240,rotate around x=45]
            \draw[dotted] (0,0,-1.5+0.3)--(0,0,0);
            \tdcyl{0}{0}{-1.5}{0.05}{0.3}
            \node at (0,0,-1.5+0.3/2) (l2) {};
            \draw[blue,->,thick] (0,0,-1.5)--(0,0,-2.1)node[anchor=south]{$\bm{z}_{2,i}\equiv\bm{w}_i$};
            \draw[thick,blue!50,->] (0,0,0-1.5)--(0.65,0,-1.5)node[anchor=north]{$\bm{y}_{2,i}$};
        	\draw[thick,blue!50,->] (0,0,0-1.5)--(0,0.5,-1.5)node[anchor=south]{$\bm{x}_{2,i}$};
        \end{scope}
        
        \tdplotdefinepoints(0,0,0)(0.7348469229,-.4242640686,-.8485281372)(1.102270384,-.4242640686,0.2121320343)
        \tdplotdrawpolytopearc[thick,dred]{0.75}{anchor=east}{\color{dred}$\alpha_1$}
        \tdplotdefinepoints(0,0,0)(1.102270384,-.4242640686,0.2121320343)(0.7076663070,-.4166905490,1.955908328)
        \tdplotdrawpolytopearc[thick,dred]{0.85}{anchor=east}{\color{dred}$\alpha_2$}
        
        \tdplotdrawarc[dgreen]{(0,0,-1.5)}{0.5}{90}{240+90}{anchor=north}{\color{dgreen}$\eta_i$}
        
        \begin{pgfonlayer}{background}
            \draw[line width=1em] (l1)to[in=-95,out=135](l2);
            \draw[line width=1em] (l2)to[in=-155,out=85](l3);
        \end{pgfonlayer}

        \begin{scope}[rotate around z=240, rotate around x=45, canvas is xy plane at z=-1.75]
            \draw [thick, safran, ->] (0.1,0,-1.75) arc (0:350:0.1)node[anchor=north,right=0.5em]{$\theta_i$};
        \end{scope}

    \end{tikzpicture}

%% file: TikZ/var_dis.tex
\def\ScaleFactor{1}

\begin{tikzpicture}[>=latex',every text node part/.style={align=center}]
	\draw[->] (0.5*\ScaleFactor,0*\ScaleFactor)--(7*\ScaleFactor,0*\ScaleFactor)node[midway,below]{space of parameters}node[right]{$\bm{U}$};
	\draw[->] (0*\ScaleFactor,0.5*\ScaleFactor)--(0*\ScaleFactor,5*\ScaleFactor)node[midway,above,sloped]{space of variables}node[above]{$\bm{X}$};
	
    
    \draw[] (0.5*\ScaleFactor,3*\ScaleFactor) to[in=120,out=85] (2*\ScaleFactor,4.5*\ScaleFactor) to[in=180,out=-60] (3*\ScaleFactor,2.5*\ScaleFactor) to[in=180,out=0] (4*\ScaleFactor,3.5*\ScaleFactor) to[in=180,out=0] (4.5*\ScaleFactor,3*\ScaleFactor) to[in=-90,out=0] (5.5*\ScaleFactor,5*\ScaleFactor);
    \draw[] (0.5*\ScaleFactor,3*\ScaleFactor) to[in=160,out=-95] (2*\ScaleFactor,1*\ScaleFactor) to[in=180,out=-20] (3*\ScaleFactor,2.5*\ScaleFactor) to[in=180,out=0] (3.75*\ScaleFactor,2.25*\ScaleFactor) to[in=180,out=0] (4.5*\ScaleFactor,3*\ScaleFactor) to[in=180,out=0] (7*\ScaleFactor,2.5*\ScaleFactor);
    \draw[densely dotted] (0.5*\ScaleFactor,3*\ScaleFactor) to[in=180,out=-80] (3*\ScaleFactor,2.5*\ScaleFactor) to[in=160,out=0] (4.5*\ScaleFactor,3*\ScaleFactor) to[in=140,out=-20] (5,1.5) to[in=180,out=-20] (7*\ScaleFactor,2*\ScaleFactor);

    \foreach \i/\j in{0.5/3,3/2.5,4.5/3}{
    	\draw[draw=none,fill=red!75,fill opacity=0.35,text opacity=1] (\i*\ScaleFactor-0.075*\ScaleFactor,\j*\ScaleFactor-0.075*\ScaleFactor) rectangle (\i*\ScaleFactor+0.075*\ScaleFactor,\j*\ScaleFactor+0.075*\ScaleFactor);
    	\draw[dashed,red] (\i*\ScaleFactor,\j*\ScaleFactor)--(\i*\ScaleFactor,0);
    	\draw[draw=none,fill=red] (\i*\ScaleFactor,0*\ScaleFactor) circle (0.05*\ScaleFactor);
    }
    
    \draw[draw=none,fill=orange!75,fill opacity=0.35,text opacity=1] (5.425*\ScaleFactor,4.35*\ScaleFactor) rectangle (5.575*\ScaleFactor,5*\ScaleFactor)node[right,orange,midway]{$\to\infty$};
    \draw[dashed,orange] (5.5*\ScaleFactor,5*\ScaleFactor)--(5.5*\ScaleFactor,0);
    \draw[draw=none,fill=orange] (5.5*\ScaleFactor,0*\ScaleFactor) circle (0.05*\ScaleFactor);
    
    \draw[line width=0.05em] (6.5*\ScaleFactor,3.5*\ScaleFactor)--(6.515*\ScaleFactor,3.5*\ScaleFactor);
    \draw[draw=none,fill=blue!75,fill opacity=0.35,text opacity=1] (6.5*\ScaleFactor-0.075*\ScaleFactor,3.5*\ScaleFactor-0.075*\ScaleFactor) rectangle (6.5*\ScaleFactor+0.075*\ScaleFactor,3.5*\ScaleFactor+0.075*\ScaleFactor);
    \draw[dashed,blue] (6.5*\ScaleFactor,3.45*\ScaleFactor)--(6.5*\ScaleFactor,0);
    \draw[draw=none,fill=blue] (6.5*\ScaleFactor,0*\ScaleFactor) circle (0.05*\ScaleFactor);
    
    \draw[line width=0.5em,draw=black!75,fill=none,draw opacity=0.25,text opacity=1] (0.65*\ScaleFactor,3.6*\ScaleFactor)to[in=115,out=70] (2*\ScaleFactor,4.5*\ScaleFactor)to[in=120,out=-60] (2.5*\ScaleFactor,2.75*\ScaleFactor) ;
    
    \begin{pgfonlayer}{background}
    	\draw[draw=none,fill=black!30,fill opacity=0.25,text opacity=1] (2.5*\ScaleFactor,2.75*\ScaleFactor)--(2.5*\ScaleFactor,0*\ScaleFactor)--(0.65*\ScaleFactor,0*\ScaleFactor)node[midway,above]{\tiny numerically\\[-1.25ex]\tiny stable interval}--(0.65*\ScaleFactor,3.6*\ScaleFactor) to[in=115,out=70] (2*\ScaleFactor,4.5*\ScaleFactor)to[in=120,out=-60] (2.5*\ScaleFactor,2.75*\ScaleFactor);
    \end{pgfonlayer}

    \draw[] (9.00*\ScaleFactor,4.15*\ScaleFactor)--(9.30*\ScaleFactor,4.15*\ScaleFactor)node[right=0.25em]{$\mathcal{V}(\mathcal{I})$};
    \draw[densely dotted] (9.00*\ScaleFactor,3.5*\ScaleFactor)--(9.30*\ScaleFactor,3.5*\ScaleFactor)node[right=0.25em]{$\mathcal{V}\left(\bra{\det{\left(\pdv{\bm{S}}{\bm{X}}\right)}}\right)$};
    \draw[draw=none,fill=red] (9.15*\ScaleFactor,2.15*\ScaleFactor) circle (0.05*\ScaleFactor)node[right=0.5em]{$\mathcal{W}_c$};
    \draw[draw=none,fill=orange] (9.15*\ScaleFactor,1.5*\ScaleFactor) circle (0.05*\ScaleFactor)node[right=0.5em]{$\mathcal{W}_\infty$};
    \draw[draw=none,fill=blue] (9.15*\ScaleFactor,0.85*\ScaleFactor) circle (0.05*\ScaleFactor)node[right=0.5em]{$\mathcal{W}_{\mathrm{sd}}$};
    \node[anchor=west] at (10.1*\ScaleFactor,1.5*\ScaleFactor) {$\left.\begin{aligned}\\[5.5ex]\\\end{aligned}\right\}\mathcal{W}_D$};
\end{tikzpicture}

%% file: TikZ/pa_co_st1_EN_overview.tex
    \begin{tikzpicture}[]
        \begin{pgfonlayer}{background}
            \node at (0,0) (pic) {\includegraphics[scale=0.575]{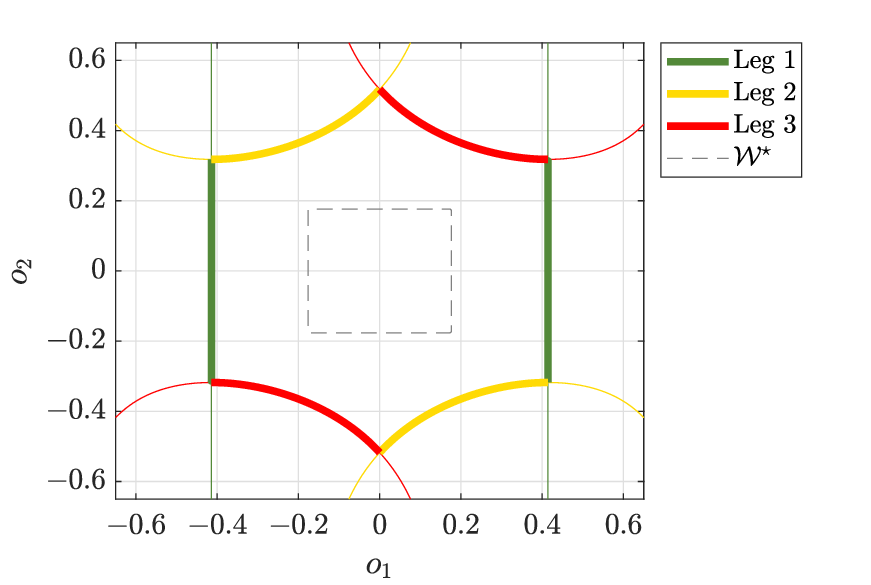}};
        \end{pgfonlayer}

        \node[draw=black,line width=0.15em,below of=pic,below=-2.5em,xshift=-17em] (s1_1) {\includegraphics[scale=0.15]{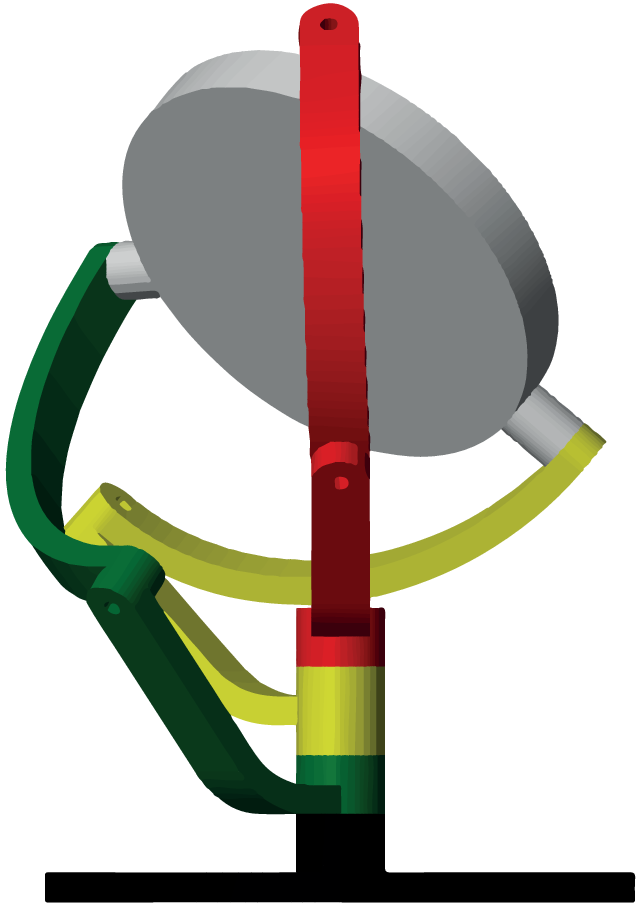}};
        \draw[black,very thick,->] (s1_1.east)to[in=-140,out=0](-1.1,-1.15);
        
        \node[draw=black,line width=0.15em,above of=pic,above=-1.25em,xshift=-17em] (s1_2) {\includegraphics[scale=0.15]{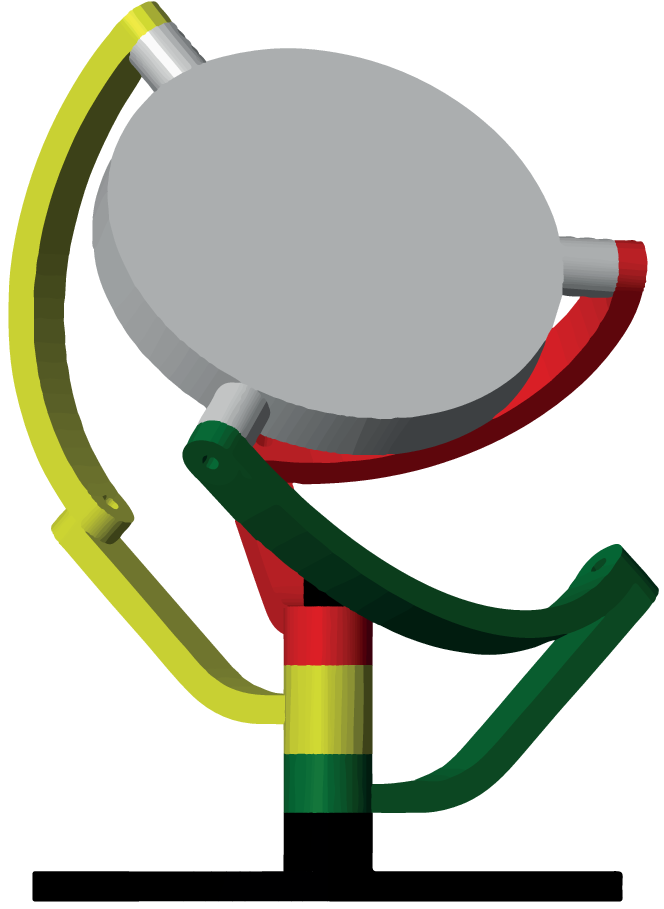}};
        \draw[black,very thick,->] (s1_2.east)to[in=140,out=0](-1.1,1.55);
        
        \node[draw=black,line width=0.15em,below of=pic,above=-4em,xshift=12em] (s1_3) {\includegraphics[scale=0.15]{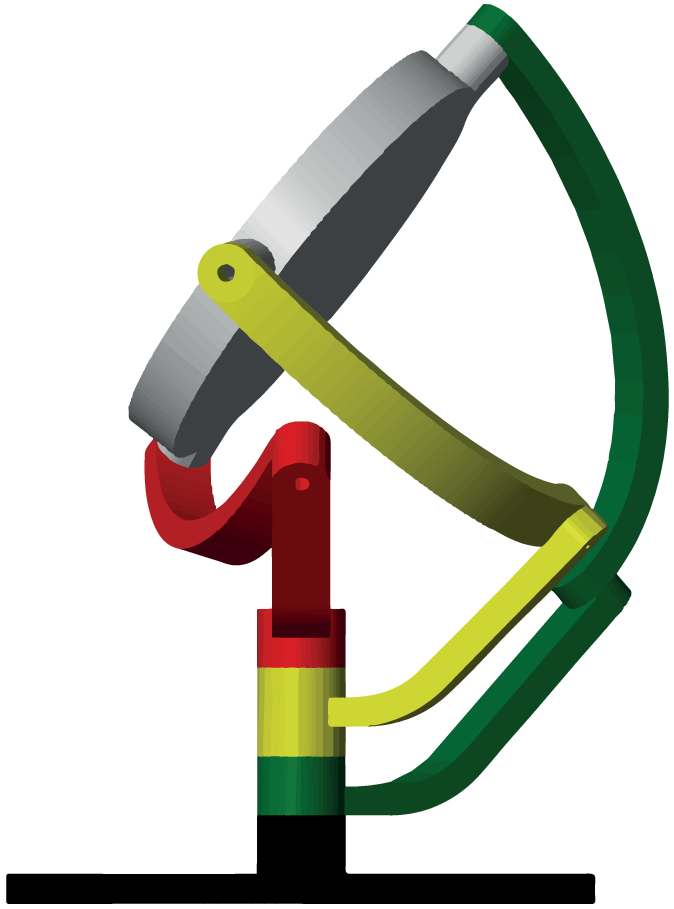}};
        \draw[black,very thick,->] (s1_3.west)to[in=0,out=-180](1.115,0.25);
    \end{tikzpicture}

%% file: TikZ/utilisation_kanto_interval.tex
	\def\ScaleFactor{0.9}
	\begin{tikzpicture}[>=latex',every text node part/.style={align=center}]
		\draw[draw=none,fill=bleu_identitaire!10,rounded corners] (-2.5*\ScaleFactor,0.5*\ScaleFactor) rectangle (2.5*\ScaleFactor,2.5*\ScaleFactor);
		\draw[draw=none] (2.5*\ScaleFactor,0.5*\ScaleFactor)--(2.5*\ScaleFactor,2.5*\ScaleFactor)node[midway,right,bleu_identitaire]{$\mathcal{W}^\star$};
		\draw[draw=none,fill=dred!10,rounded corners] (-2.5*\ScaleFactor,-0.5*\ScaleFactor) rectangle (2.5*\ScaleFactor,-2.5*\ScaleFactor);
		\draw[draw=none](2.5*\ScaleFactor,-0.5*\ScaleFactor) rectangle (2.5*\ScaleFactor,-2.5*\ScaleFactor)node[midway,right,dred]{$\mathcal{Q}^\star$};
		
		\begin{pgfonlayer}{foreground}
		\draw[black,fill=black] (-1.5*\ScaleFactor,-1.5*\ScaleFactor)node(j0){}node[below]{$\bm{j}_0$} circle (0.05*\ScaleFactor);
		\draw[black,fill=black] (-1.5*\ScaleFactor,1.5*\ScaleFactor)node(o0){}node[above]{$\bm{o}_0$} circle (0.05*\ScaleFactor);
		\end{pgfonlayer}
		\draw[->,line width=0.05em] (j0)to[in=-124,out=120](o0)node[fill=white,midway,left=4.5em]{\scriptsize known\\[-1ex]\scriptsize solution};
		
		
		\draw[dashdotted,thin,fill=black!15,fill opacity=0.25] (j0) circle (2.3em);
		\draw[black,fill=black] (-0.85*\ScaleFactor,-1.25*\ScaleFactor)node(j1){}node[below]{$\bm{j}_1$} circle (0.05*\ScaleFactor);
		\draw[->,line width=0.15em] (j0)--(j1);		
		\draw[dotted,thin] (j0)--($(j0)+(0,-1.45*\ScaleFactor)$);
		\draw[dotted,thin] ($(j0)+(2.3em,0)$)--($(j0)+(2.3em,0)+(0,-1.45*\ScaleFactor)$);
		\draw[<->,thick] ($(j0)+(0,-1.4*\ScaleFactor)$)--($(j0)+(2.3em,0)+(0,-1.4*\ScaleFactor)$)node[right]{\scriptsize system precision};

		\draw[densely dashdotted,thin,fill=black!75,fill opacity=0.15] (-0.85*\ScaleFactor,1.25*\ScaleFactor) circle (0.125*\ScaleFactor);
		\draw[black,fill=black] (-0.85*\ScaleFactor,1.25*\ScaleFactor)node(o1){}node[above=0.2em]{$\bm{o}_1$} circle (0.05*\ScaleFactor);
		\draw[->,line width=0.05em] (j1)to[in=-124,out=120](o1)node[fill=white,midway,left=1.25em]{\scriptsize certified\\[-1ex]\scriptsize Newton};

		\draw[dashdotted,thin,fill=black!15,fill opacity=0.25] (o0) circle (2.3em);

		\draw[->,line width=0.15em] (o0)--(o1);

		\foreach \i/\j/\k[count=\xi from 2] in{-0.25/1.55/2}{
			\draw[black,fill=black] (\i*\ScaleFactor,-\j*\ScaleFactor)node(j\xi){}node[below]{$\bm{j}_{\k}$} circle (0.05*\ScaleFactor);		
			\draw[black,fill=black] (\i*\ScaleFactor,\j*\ScaleFactor)node(o\xi){}node[above]{$\bm{o}_{\k}$} circle (0.05*\ScaleFactor);
		}

		\foreach \i/\j/\k[count=\xi from 3] in{1.15/1.55/k,1.85/1.35/k+1}{
			\draw[black,fill=black] (\i*\ScaleFactor,-\j*\ScaleFactor)node(j\xi){}node[below]{$\bm{j}_{\k}$} circle (0.05*\ScaleFactor);
			\draw[black,fill=black] (\i*\ScaleFactor,\j*\ScaleFactor)node(o\xi){}node[above]{$\bm{o}_{\k}$} circle (0.05*\ScaleFactor);
		}
		
		\draw[->,line width=0.15em] (j1)--(j2);
		\draw[->,line width=0.15em,dashed] (j2)--(j3);
		\draw[->,line width=0.15em] (j3)--(j4);
		\draw[->,line width=0.15em] (o1)--(o2);
		\draw[->,line width=0.15em,dashed] (o2)--(o3);
		\draw[->,line width=0.15em] (o3)--(o4);
	\end{tikzpicture}